%% file: main.tex
\def\year{2020}\relax
\documentclass[letterpaper]{article} % DO NOT CHANGE THIS
\usepackage{aaai20}  % DO NOT CHANGE THIS
\usepackage{times}  % DO NOT CHANGE THIS
\usepackage{helvet} % DO NOT CHANGE THIS
\usepackage{courier}  % DO NOT CHANGE THIS
\usepackage[hyphens]{url}  % DO NOT CHANGE THIS
\usepackage{graphicx} % DO NOT CHANGE THIS
\usepackage{graphicx}  % DO NOT CHANGE THIS
\usepackage{framed} %
\usepackage[table, svgnames]{xcolor}
\usepackage{lipsum} %
\usepackage[defaultlines=2,all]{nowidow}

\colorlet{barcolour}{Blue!50!Gainsboro}
\renewenvironment{leftbar}{%
	\MakeFramed {\advance\hsize-\width \FrameRestore}\noindent\hspace{-\fontdimen2\font}}%
{\endMakeFramed}

\usepackage{soul}
\definecolor{lightblue}{rgb}{.68,.85,0.90}
\sethlcolor{lightblue}
\newtheorem{definition}{Definition}
\usepackage{comment}
\usepackage{url}
\usepackage{booktabs}
\usepackage{makecell}
\usepackage{multirow}
\usepackage{bm}
\usepackage{csquotes}
\usepackage{enumitem}
\usepackage[brackets]{schemata}
\usepackage{etoolbox}
\usepackage{mathrsfs, amssymb}
\usepackage{mathtools}
\usepackage{pifont}
\usepackage{footnote}
\usepackage{footmisc}
\usepackage{xcolor}
\makesavenoteenv{tabular}
\makesavenoteenv{table}
\usepackage{lipsum,booktabs}
\usepackage[caption=false]{subfig}
\usepackage{algorithm,algpseudocode}

\title{MILE: A Multi-Level Framework for  Scalable Graph Embedding}
\author{
Jiongqian Liang \thanks{Equal contribution}\thanks{Now at Google},
Saket Gurukar \textsuperscript{*},
Srinivasan Parthasarathy \\
Department of Computer Science and Engineering, \\ 
The Ohio State University \\
liang.420@osu.edu, gurukar.1@osu.edu, srini@cse.ohio-state.edu
}

\newif\ifext % extended version or not.
\extfalse

\begin{document}
\maketitle

\begin{abstract}
	\input{abstract.tex}
\end{abstract}

\section{Introduction}
\input{introduction.tex}

\section{Related Work}
\input{relatedWork.tex}

\section{Problem Formulation}
\input{problemFormulation.tex}

\section{Methodology}
\input{methodology.tex}

% \vspace{5em}
\section{Experiments and Analysis}
\input{experiments.tex}

\section{Conclusion}
\input{conclusion.tex}

\bibliographystyle{aaai}
\bibliography{iclr_citations}

\end{document}

\begin{comment}
% Comment out already
\subsection{Graph Coarsening}
\label{appendix:graph_coarsening}
In this phase, the input graph $\mathcal{G}$ (or $\mathcal{G}_0$) is repeatedly coarsened into a series smaller graphs 
$\mathcal{G}_1$, $\mathcal{G}_2$, $...$, $\mathcal{G}_m$ such that $|V_0| > |V_1| > ... > |V_m|$.
\begin{figure}[!htb]
	% \includegraphics[width=0.45\textwidth]{./Figures/toy_example.pdf}
	% \includegraphics[width=0.45\textwidth]{./Figures/toy_example.pdf}
	\subfloat[Using SEM and NHEM for graph coarsening]{\label{fig:toy1}\includegraphics[width=0.55\textwidth]{./Figures/toy1.pdf}} \hspace{2mm}
	\subfloat[Adjacency matrix and matching matrix]{\label{fig:toy2}\includegraphics[width=0.38\textwidth]{./Figures/toy2_V.pdf}}
	%\vspace{-2mm}
	\caption[xxx]{\small
		Toy example for illustrating graph coarsening\footnotemark.
		(a) shows the process of applying Structural Equivalence Matching (SEM) and Normalized Heavy Edge Matching (NHEM) for graph coarsening.
		(b) presents the adjacency matrix $A_0$ of the input graph, the matching matrix $M_{0,1}$ corresponding to the SEM and NHEM matchings, and the derivation of the adjacency matrix $A_1$ of the coarsened graph using Eq.~\ref{eq:coarsen}.} 
	\label{fig:toy}
\end{figure}

\end{comment}

% \input{drilldown.tex}

\section{Time Complexity}
\label{appendix:time}
	
 It is non-trivial to derive the exact time complexity of MILE as it is dependent on the graph structure, the chosen base embedding method, and the convergence rate of the GCN model training. Here, we provide a rough estimation of the time complexity.  For simplicity, we assume the number of vertices and the number of edges are reduced by factor $\alpha$ and $\beta$ respectively at each step of coarsening 
($\alpha>1.0$ and $\beta>1.0$),
i.e., $V_i = \frac{1}{\alpha}{V_{i-1}}$ and $E_i = \frac{1}{\beta}{E_{i-1}}$. (we found $\alpha$ and $\beta$ in range $[1.5, 2.0]$, empirically).  With $m$ levels of coarsening, the coarsening complexity is approximately $O((1- 1/\beta \textsuperscript{m})/(1- 1/\beta) \times E))$ and since $1/\beta\textsuperscript{m}$ is small, the complexity reduces to $O(\frac{\beta}{\beta-1} \times E)$.
For the base embedding phase, if the embedding algorithm has time complexity of $T(V, E)$, the complexity of the base embedding phase is
$T(\frac{V}{{\alpha}^m}, \frac{E}{{\beta}^m})$.
For the refinement phase, the time complexity can be divided into two parts, i.e. the GCN model training and the embedding inference applying the GCN model.
The former has similar complexity as the original GCN and can be denoted as $O(k_1*\frac{E}{\beta ^m})$~\cite{GCN}, where $k_1$ is a small constant related to embedding dimensionality and the number of training epochs. 
%Note that the model training happends at the coarsest graph so it is supposed to very efficient. 
The embedding inference part is simply sparse matrix multiplication using Eq. %~\ref{eq:layer}
$H^{(k)}(X, A) = \sigma \left( \tilde{D}^{-\frac{1}{2}} \tilde{A} \tilde{D}^{-\frac{1}{2}} H^{(k-1)}(X, A) \Theta^{(k)} \right)$
with time complexity $O(k_2*E_i)$
when refining the embeddings on graph $\mathcal{G}_i$, where $k_2$ is an even smaller
constant ($k_2 < k_1$). As a result, the time complexity of the whole refinement phase is 
$O(k_1*\frac{E}{\beta ^m} +  k_2*(E + \frac{E}{\beta ^1} ... + \frac{E}{\beta ^{m-1}})) \approx O(k_3 * E)$ where $k_3$ is a small constant.
%Note that the model training happens at the coarsest level, so $E_m$ is only a small fraction of the number of edges in the original graph $E$.
%After the refinement model being trained, the time complexity of applying the refinement model between $\mathcal{G}_i$ and $\mathcal{G}_i-1$ is 
%$O(E_i)$~\cite{GCN}. As a result, the whole refinement phase has time complexity $O(k*E)$.

Overall, for an embedding algorithm of time complexity $T(V, E)$, the MILE framework can reduce it to be 
$T(\frac{V}{{\alpha}^m}, \frac{E}{{\beta}^m}) + O(k*E)$. This is a significant improvement considering $T(V, E)$ is usually very large. 
The reduction in time complexity is attributed to the fact that we run the embedding learning and refinement model training at the coarsest graph. 
In addition, the overhead introduced by
the coarsening phase and recursive embedding refinement is relatively small (linear to the number of edges $E$). 
Note that the constant factor $k$ in the complexity term is usually small and we empirically found it to be in the scale of tens. 
Because of this, even when the complexity of the original embedding algorithm is linear to $E$, our MILE framework
could still potentially speed up the embedding process because the complexity of MILE contains a smaller constant factor $k$. %(see Sec.~\ref{sec:exp1} for the experiment of applying MILE on LINE). 

Furthermore,
it is worth noting that many of the existing embedding strategies 
involve hyperparameters tunning for the best performance,
especially for those methods based on neural networks (e.g., DeepWalk, Node2Vec, etc.). 
%Another advantage of MILE particularly when applied to base embedding strategies that rely on a neural-network architecture (e.g. DeepWalk, Node2Vec…)
%  is the fact that such base embeddings need to be tuned (to find the best set of hyperparameter values) for a given dataset.  
This in turn requires the algorithm to be run repeatedly -- hence any savings in runtime by applying MILE are magnified across multiple runs of the algorithm with different hyper-parameter settings.

{}

\section{MILE Drilldown: Discussion on reusing $\Theta^{(k)}$  across all levels}

 Similar to GCN, $\Theta^{(k)}$  is a matrix of filter parameters and is of size $d*d$ (where $d$ is the embedding dimensionality). Eq. %\ref{eq:layer} 
 $H^{(k)}(X, A) = \sigma \left( \tilde{D}^{-\frac{1}{2}} \tilde{A} \tilde{D}^{-\frac{1}{2}} H^{(k-1)}(X, A) \Theta^{(k)} \right)$
 in this paper defines how the embeddings are propagated during embedding refinements, parameterized by $\Theta^{(k)}$ . Intuitively,  $\Theta^{(k)}$  defines how different embedding dimensions interact with each other during the embedding propagation. This interaction is dependent on graph structure and base embedding method, which can be learned from the coarsest level.  Ideally, we would like to learn this parameter $\Theta^{(k)}$  on every two consecutive levels. But this is not practical since this could be expensive as the graph get more fine-grained (and defeat our purpose of scaling up graph embedding). This trick of ``sharing'' parameters across different levels is the trade-off between efficiency and effectiveness. To some extent, it is similar to the original GCN \cite{GCN}, where the authors share the same filter parameters $\Theta^{(k)}$  over the whole graph (as opposed to using different $\Theta^{(k)}$  for different nodes; see Eq (6) and (7) in\cite{GCN}). Moreover, we empirically found this works good enough and much more efficient. Table \ref{tab2} shows that if we do not share $\Theta^{(k)}$  values and use random values for $\Theta^{(k)}$  during refinements, the quality of embedding is much worse (see baseline MILE-untr). 
 
 \section{MILE Drilldown: Discussion on  choice of embedding methods}
 We wish to point out that we chose the base embedding methods as they are either recently proposed NetMF (introduced in 2018) or are widely used (DeepWalk, Node2Vec, LINE). By showing the performance gain of using MILE on top of these methods, we want to ensure the contribution of this work is of broad interest to the community.  We also want to reiterate that these methods are quite different in nature: 
 \begin{itemize}

\item DeepWalk (DW) and Node2vec (N2V) rely on the use of random walks for latent representation of features.
\item  LINE learns an embedding that directly optimizes a carefully constructed objective function that preserves both first/second order proximity among nodes in the embedding space.
\item  GraRep constructs multiple objective matrices based on high orders of random walk laplacians, factories each objective matrix to generate embeddings and then concatenates the generated embeddings to form final embedding.
\item  NetMF constructs an objective matrix based on random walk Laplacian and factorizes the objective matrix in order to generate the embeddings.
  \end{itemize}
 
 Indeed NetMF \cite{NETMF, levy2014neural} with an appropriately constructed objective matrix has been shown to {\it approximate} DW, N2V and LINE allowing such be conducting implicit matrix factorization of {\it approximated} matrices. There are limitations to such approximations (shown in a related context by \cite{arora2016tacl}) - the most important one is the requirement of a sufficiently large embedding dimensionality.  Additionally, we note that while unification is possible under such a scenario, the methods based on matrix factorization are quite different from the original methods and do place a much larger premium on space (memory consumption) - in fact this is observed by the fact we are unable to run NetMF and GraRep in many cases without incorporating them within MILE.

 \section{MILE Drilldown: Discussion on  extending MILE to directed graphs}
 
 Note that as pointed out by \cite{chung2005laplacians}, one can construct random-walk Laplacians for a directed graph thus incorporating approaches like NetMF to accommodate such solutions.  Another simple solution is to symmetrize the graph while accounting for directionality. Once the graph is symmetrized, any of the embedding strategies we discuss can be employed within the MILE framework (including the coarsening technique). There are many ideas for symmetrization of directed graphs (see for example work described by \cite{gleich2006hierarchical} or \cite{satuluri2011symmetrizations}.

 \section{MILE Drilldown: Discussion on  effectiveness of SEM}
 
The effectiveness of structurally equivalent matching (SEM) is highly dependent on graph structure but in general 5\% - 20\% of nodes are structurally equivalent (most of which are low-degree nodes). For example, during the first level of coarsening, YouTube has 172,906 nodes (or 86,453 pairs) out of 1,134,890 nodes that are found to be SEM (around 15\%); Yelp has 875,236 nodes (or 437,618 pairs) out of 8,938,630 nodes are SEM (around 10\%). In fact, more nodes are involved in SEM as SEM is run iteratively at each coarsening level.

% \vspace{1em}
% \clearpage

\bibliography{iclr_citations}
\bibliographystyle{abbrv}

%% file: abstract.tex
 Recently there has been a surge of interest in designing graph embedding methods. Few, if any, can scale to a large-sized 
graph with millions of nodes due to both computational complexity and memory requirements.
In this paper, we relax this limitation by introducing the MultI-Level Embedding (MILE) framework -- a generic methodology allowing contemporary graph embedding methods to scale to large graphs.
MILE repeatedly coarsens the graph into smaller ones using a hybrid matching technique 
to maintain the backbone structure of the graph.
It then applies existing embedding methods on the coarsest graph and 
refines the embeddings to the original graph through a graph convolution neural network that it learns.
%refines the embeddings to the original graph through a novel, graph convolution procedure, it learns.
% using the graph convolution network.
% The third and refines the embeddings to the original graph. 
%In order to capture more holistic features of the graph and boost the speed of graph embedding, 
The proposed MILE framework is agnostic to the underlying graph embedding techniques and can be applied
to many existing graph embedding methods without modifying them.
We employ our framework on several popular graph embedding techniques and conduct embedding for real-world graphs.
Experimental results on five large-scale datasets demonstrate that MILE significantly boosts the speed (order of magnitude)
of graph embedding while generating embeddings of better quality, for
% efficiency and efficacy of graph embedding. 
%Experimental results on various datasets with different base embedding techniques demonstrate the power of MILE for on both improving the efficiency and efficacy of graph embedding. 
%On the larger datasets, we achieve speedup as large as 11X along with $10\%$ improvement on the Macro-F1 for
the task of node classification.
MILE can comfortably scale to a graph with 9 million nodes and 40 million edges, 
on which existing methods run out of memory or take too long to compute on a modern workstation.
Our code and data are publicly available with detailed instructions for adding new base embedding methods: \url{https://github.com/jiongqian/MILE}. 

%% file: introduction.tex
%Embedding strategies are an increasingly popular approach to non-linear dimensionality reduction and find significant use

In recent years, {\it network embedding} has attracted much interest due to its broad applicability for a range of tasks such as social influence prediction \cite{qiu2018deepinf}, network role discovery\cite{rossi2014role}, political perspective detection \cite{li2019encoding}, and social recommender systems \cite{wu2018socialgcn}.
\begin{comment}
\textcolor{blue}{COMMENTED OUT\\
role discovery~\cite{rossi2014role}, query pattern matching~\cite{Zou12}, 
 visualization~\cite{VDM08},
% ~\cite{H12}
including database cleaning~\cite{Chu16},
% query pattern matching~\cite{Zou12}, 
exploratory data analysis~\cite{Zhang12},
distance oracles~\cite{Z13,Qi13}, 
classification~\cite{DEEPWALK}
and link prediction~\cite{SDNE}
}
\end{comment}
%As a result of its broad applicability to various domains, research in this area has seen an explosive growth of interest.
While these new embedding methods often offer a competitive qualitative advantage over traditional approaches,
many of them do not scale to large datasets (e.g., graphs with over 1 million nodes) since they are computationally expensive and often memory intensive. 
For example, random-walk-based embedding techniques such as DeepWalk~\cite{DEEPWALK} and Node2Vec~\cite{NODE2VEC}, 
require a large amount of CPU time to generate a sufficient number of walks and train the embedding model. 
It takes over $10$ hours for a single machine to achieve quality embeddings on a graph with a million nodes using such methods. 
\begin{comment}
\textcolor{blue}{COMMENTED OUT\\
As another example, embedding methods based on matrix factorization, including GraRep~\cite{GRAREP} and NetMF~\cite{NETMF}, requires constructing an enormous objective matrix (usually much denser than adjacency matrix), on which matrix factorization is performed. Even a medium-size graph with 100K nodes can easily require hundreds of GB of memory using those methods.
On the other hand, many graph datasets in the real world tend to be large-scale with millions or even billions of nodes.
}
\end{comment}
To the best of our knowledge, none of the existing efforts examines how to scale up graph embedding in a \textbf{generic} way. We make the first attempt to close this gap. We are also interested in the related question of whether the quality of such embeddings can be improved along the way. Specifically, we ask:
\begin{enumerate}

\item  Can we scale up the existing embedding techniques in an agnostic manner so that they can be directly applied to larger datasets?

\item  Can the quality of such embedding methods be strengthened by incorporating the holistic view of the graph? 
%\extvspace{-4mm}
\end{enumerate}

To tackle these problems, we propose a \underline{M}ult\underline{I}-\underline{L}evel \underline{E}mbedding (MILE) framework for graph embedding. Our approach relies on a three-step process: {\bf first}, we repeatedly coarsen the original graph into smaller ones
by employing a hybrid matching strategy; {\bf second}, we compute the embeddings on the coarsest graph using an existing  
embedding technique - 
note that graph embedding on the coarsest graph is inexpensive to compute and utilizes far less memory,
and moreover intuitively can capture the global structure of the original graph~\cite{METIS,MLR-MCL}; 
and {\bf third},
we propose a novel refinement model based on learning a graph convolution network to refine the embeddings from 
the coarsest graph to the original graph -- learning a graph convolution network allows us to attain a refinement procedure that levers the dependencies inherent to the graph structure and the embedding method of choice. 
%MILE's ability to learn a data- and embedding- sensitive refinement procedure is key to its effectiveness.
%Generating graph embeddings by leveraging hierarchical information of graph is also employed by techniques like HARP \cite{HARP}, however, in this work, we introduce the refinment model which helps us tackle the problem of scalability.
To summarize, we find:

%\smallskip
\begin{itemize}

\item  MILE is generalizable: Our MILE framework is agnostic to the underlying graph embedding techniques and treats them as black boxes.
%We report results on DeepWalk\cite{DEEPWALK}, Node2Vec\cite{NODE2VEC}, GraRep\cite{GRAREP}, and NetMF\cite{NETMF}.

%\smallskip
\item  MILE is scalable: MILE can {\it significantly improve the scalability of the embedding methods} ({\bf up to 30-fold}), by reducing the running time and memory consumption. 
%We particularity illustrate this on a large graph with over 9 million nodes and 40 million edges -- a dataset on which several baselines fail to run (out of memory) and we can compute quality embeddings in about 2.5 hours on a modern workstation . %%REDUNDANT WITH ABSTRACT

%\smallskip
\item   MILE generates high-quality embeddings: In many cases, we find that the quality of embeddings improves by levering MILE (in some cases is in excess of  10\%). 

%\smallskip
%\item   MILE's ability to learn a data- and embedding- sensitive refinement procedure is key to its effectiveness.
%Other design choices such as the hybrid coarsening strategy also enable MILE to produce quality embeddings in a scalable fashion. 

\end{itemize}

%% file: relatedWork.tex
\label{sec:related}
%\extvspace{-1mm}

%Network embedding has gained significant interest in recent years. 
% Early techniques include IsoMap~\cite{isomap} and Locally Linear Embedding (LLE)~\cite{LLE}.
\noindent 
\textbf{Network Embedding}:
Many techniques for graph or network
embedding have been proposed in recent years.
DeepWalk and Node2Vec generate truncated random walks on graphs and 
apply the Skip Gram by treating the walks as sentences~\cite{DEEPWALK,NODE2VEC}.
% As a generalized version of DeepWalk, Node2Vec uses the same model but generates paths with more flexibility on the breadth and depth of the the random walks~\cite{NODE2VEC}.
LINE learns the node embeddings by preserving the first-order and second-order proximities~\cite{LINE}.
Following LINE, SDNE leverages deep neural networks to capture the highly non-linear structure~\cite{SDNE}.
Other methods construct a particular objective matrix and use matrix factorization techniques to generate embeddings,
e.g., GraRep~\cite{GRAREP} and NetMF~\cite{NETMF}.
This also led to the proliferation of network embedding methods for information-rich graphs, 
including heterogeneous information networks~\cite{chang2015heterogeneous,dong2017metapath2vec} and
attributed graphs~\cite{triparty,SEANO,GCN,TADW}. %

There are very few efforts, focusing on the scalability of network embedding~\cite{yang2017fast,AANE}.
First, such efforts are specific to a particular embedding strategy and do not generalize. Second, the scalability of such efforts is  limited to moderately sized datasets. Finally, and notably, these efforts at scalability are actually orthogonal to our strategy and can potentially be employed along with our efforts to afford even greater speedup. 

%\hspace{-0.3in}\schema[open]{}{
The closest work to this paper is HARP~\cite{HARP}, which proposes a hierarchical paradigm for graph embedding based on
iterative learning methods (e.g., DeepWalk and Node2Vec).
However, HARP focuses on improving the quality of embeddings by using 
the learned embeddings from the previous level
as the initialized embeddings for the next level, which introduces a huge computational overhead.
Moreover, it is not immediately obvious how a HARP like a methodology would be extended to other graph embedding techniques (e.g., GraRep and NetMF) in an agnostic manner since such an approach would necessarily require one to modify the embedding methods to preset their initialized embeddings. 
In this paper, we focus on designing a general-purpose framework to scale up embedding methods treating them as black boxes. 

\noindent
\textbf{Multi-level Community Detection}:
%\ifext
The multi-level approach has been widely studied for efficient community detection~\cite{METIS,MLR-MCL,GRACLUS,ruan2015community}.
%\else
%The multi-level approach has been widely studied for efficient community detection~\cite{METIS,MLR-MCL,GRACLUS}.
%\fi
The key idea of these multi-level algorithms is to coarsen the original graph into a much smaller one, which is
then partitioned into clusters. The partitions are then recovered from the coarse-grained graph to the original graph in a recursive manner.
While our framework shares some ideas at a conceptual level with such efforts, the objectives are distinct in that we focus on graph embeddings while these methods work on graph partitioning and community discovery.

% motivation behind all multilevel algorithms is to efficiently
% obtain a partitioning of nodes at a coarse-grained level, and then recover
% fine-grained communities from high-level clusters in a recursive manner
% of graph coarsening is popular in community discovery algorithms and forms the basis for algorithms like METIS~\cite{METIS}, MLR-MCL~\cite{MLR-MCL}, Graclus~\cite{GRACLUS} and Louvain~\cite{LOUVAIN}.

% Yang et al.~\cite{yang2017fast} proposed a model which enhances the quality of embeddings by  approximating the higher order proximity matrix. Huang et al.~\cite{AANE} proposed a model which generates embeddings for attributed networks based on the decomposition of attribute affinity matrix and the penalty of embedding difference between connected nodes. Embedding techniques~\cite{qu2017attention,zitnik2017predicting} for Multi-Layer Networks also exist in which the input graph itself has multiple layers.

%Defferrard et al.~\cite{defferrard2016convolutional}  generalize CNN to graphs using tools from Graph Signal Processing.  

%\begin{comment}~\cite{ISOMAP,  }\end{comment}

%\end{comment}

% ISOMAP: A Global Geometric Framework for Nonlinear Dimensionality Reduction
% Nonlinear Dimensionality Reduction by Locally Linear Embedding.

%% file: problemFormulation.tex
% \subsection{Problem Formulation}
Let $\mathcal{G} = (V, E)$ be the input graph % (weighted or unweighted), 
where $V$ and $E$ are respectively the node set and edge set.
Let $A$ be the adjacency matrix of the graph  %$|V|\times |V|$ 
%with each entry $A(u, v)$ denoting the weight of the edge between node $u$ and $v$. 
%Without ambiguity, we refer to the graph as $\mathcal{G}$ and $A$ interchangeably in the rest of the paper.
and we assume $\mathcal{G}$ is undirected, though our problem can be easily extended \cite{chung2005laplacians,gleich2006hierarchical,MLR-MCL} to a directed graph. Table {\ref{tab:notation} shows the table of notations.} We first define graph embedding:

%\extvspace{-1mm}
\begin{definition}{\textit{\textbf{Graph Embedding}}}
Given a graph $\mathcal{G} = (V, E)$ and a 
\ifext
pre-defined 
\fi
dimensionality $d$ ($d \ll |V|$),
the problem of graph embedding is to learn a $d$-dimension vector representation for 
each node in 
\ifext
    graph
\fi
$\mathcal{G}$ so that graph properties are best preserved.
%\extvspace{-2mm}
\end{definition}
Following this, a graph embedding method is essentially a mapping 
function $f: \mathbb{R}^{|V|\times |V|} \mapsto \mathbb{R}^{|V|\times d}$, whose input is the adjacency matrix $A$ (or $\mathcal{G}$)
and output is a lower dimension matrix.  Motivated by the fact that the majority of graph embedding methods cannot scale to large datasets, we seek to speed up 
existing graph embedding methods without sacrificing quality.
We formulate the problem as: 

\textit{ Given a graph $\mathcal{G} = (V, E)$ and a graph embedding method $f(\cdot)$, we aim to 
realize a strengthened graph embedding method $\hat{f}(\cdot)$ so that it is more scalable %computationally efficient 
than $f(\cdot)$ while generating embeddings of comparable or even better quality}.

%We refer to the process of applying $f(\cdot)$ on a graph as \textit{base embedding}, where $f(\cdot)$ is called the \textit{base embedding method}.

%% file: methodology.tex
\begin{figure}[t]
    % \vspace{-2mm}
    \begin{center}
        \includegraphics[width=0.35\textwidth]{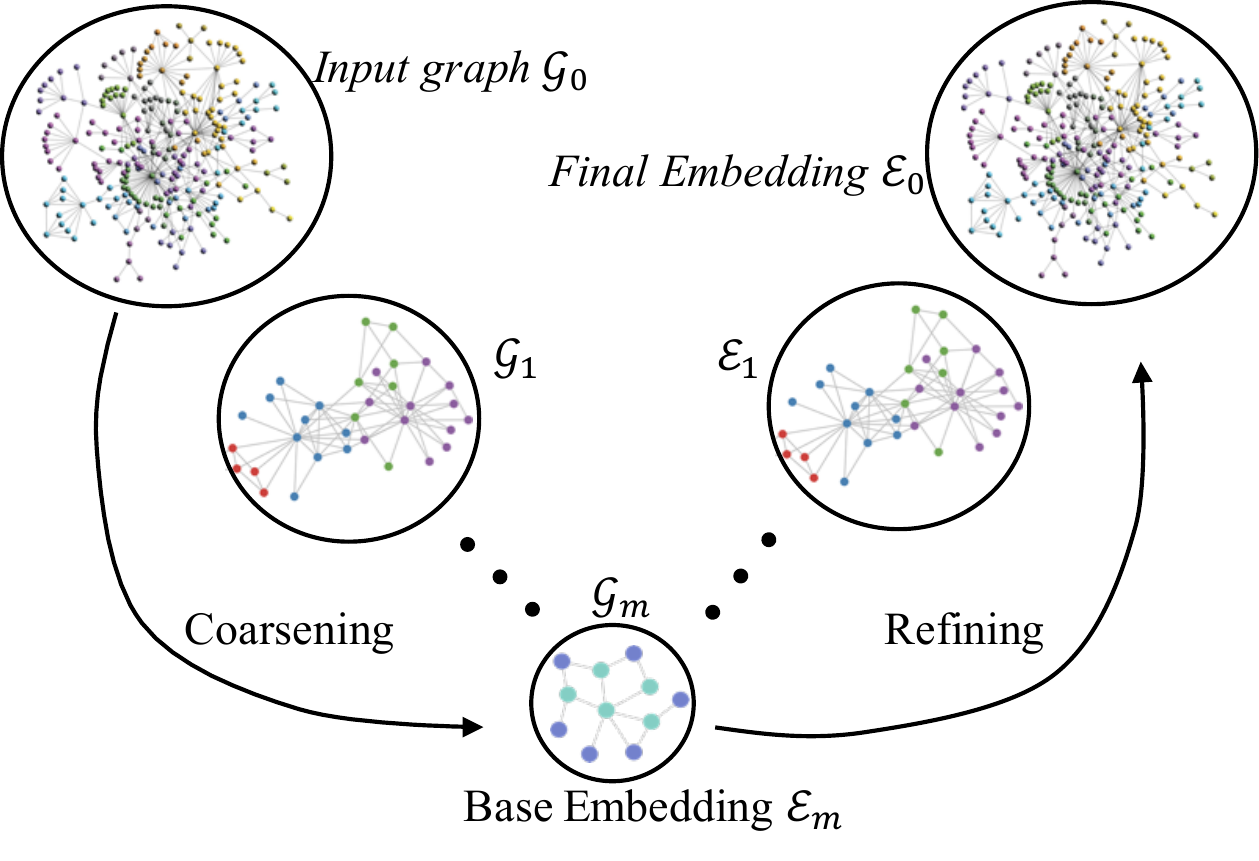}
    \end{center}
%   \vspace{-3mm}
    \caption{An overview of the multi-level embedding framework}
    \label{fig:overview}
%    \vspace{-4mm}
\end{figure}

\begin{comment}
\begin{figure*}
        \begin{center} 
    \quad\quad
    \subfloat[\small{An overview of the multi-level embedding framework.}]{
            \includegraphics[width=0.6\linewidth]{./Figures/overview.pdf}
                          \label{fig:overview}
    }        \subfloat[  Architecture of the embeddings refinement model.]{
            \includegraphics[width=0.6\linewidth]{./Figures/ref_arch.png}
              \label{fig:ref_arch}
    }
        \end{center}      
\caption{MILE framework}
%\vspace{1.5em}
\end{figure*}
\end{comment}

\begin{table}
\centering
\resizebox{0.8\columnwidth}{!}{ 
\begin{tabular}{l|l}
\toprule \hline
\textbf{Symbol} & \textbf{Definition} \\ \hline \hline
$\mathcal{G}_{i}$ &  the graph after $i$ iterations of coarsening \\ \hline
% $n_i$, $m_i$ & $\#$ vertices, \# edges in $\mathcal{G}_{i}$ \\ \hline
$V_i$, $E_i$  &  vertex set, edge set of $\mathcal{G}_{i}$ \\ \hline
$A_{i}$, $D_{i}$ &  the adjacency and degree matrix of $\mathcal{G}_{i}$ \\ \hline
% $D_{i}$ &  the degree matrix of $\mathcal{G}_{i}$ \\ \hline
$d$          &  dimensionality of the embeddings \\ \hline
$m$          &  the total number of coarsening levels \\ \hline
$f(\cdot)$ &  the base embedding method applicable on $\mathcal{G}_i$ \\ \hline
$\mathcal{E}_{i}$ &  the embeddings of nodes in $\mathcal{G}_{i}$ \\ \hline
$M_{i, i+1}$ & the matching matrix from $\mathcal{G}_{i}$ to $\mathcal{G}_{i+1}$ \\ \hline
$\mathcal{R} $($\cdot$) &  the embeddings refinement model \\ \hline
$l$          &  \# layers in the graph convolution network \\ \hline
\bottomrule
\end{tabular}
}
% \vspace{1mm}
\caption{The table of notations.}
%\extvspace{-5mm}
\label{tab:notation}
\end{table}

\begin{comment}
\end{comment}

\ifext  
\begin{figure*}
\subfloat[Using SEM and NHEM for graph coarsening]{\label{fig:toy1}\includegraphics[width=0.8\textwidth]{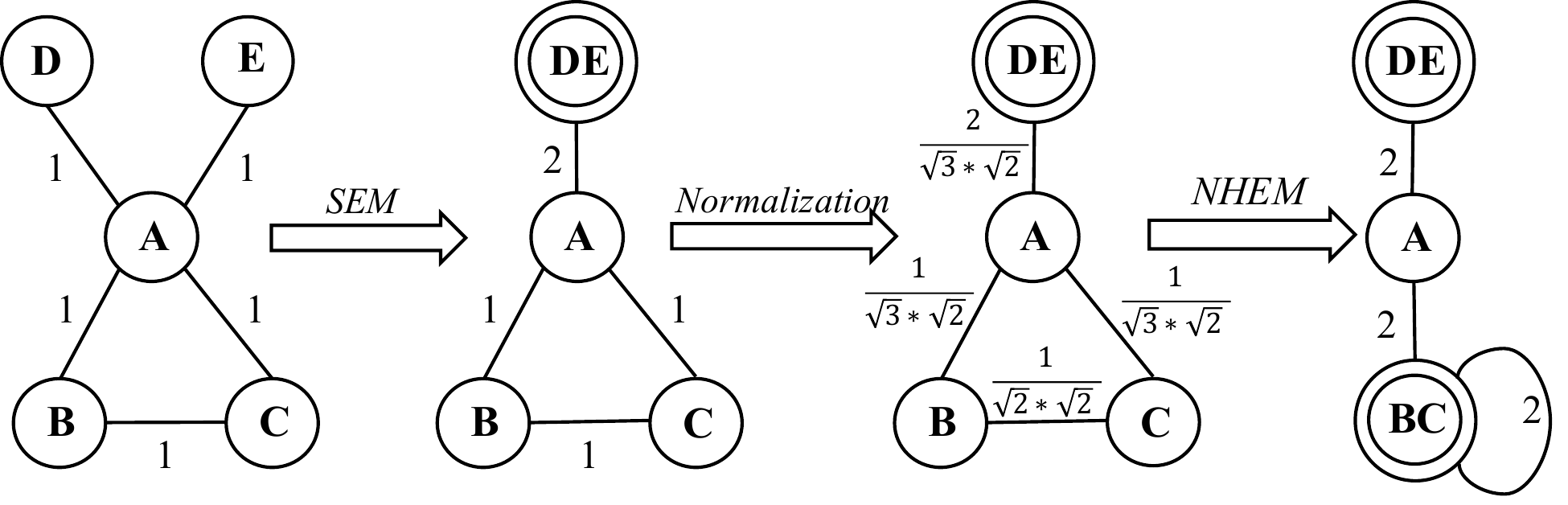}} \hspace{2mm}
\subfloat[Adjacency matrix and matching matrix]{\label{fig:toy2}\includegraphics[width=0.7\textwidth]{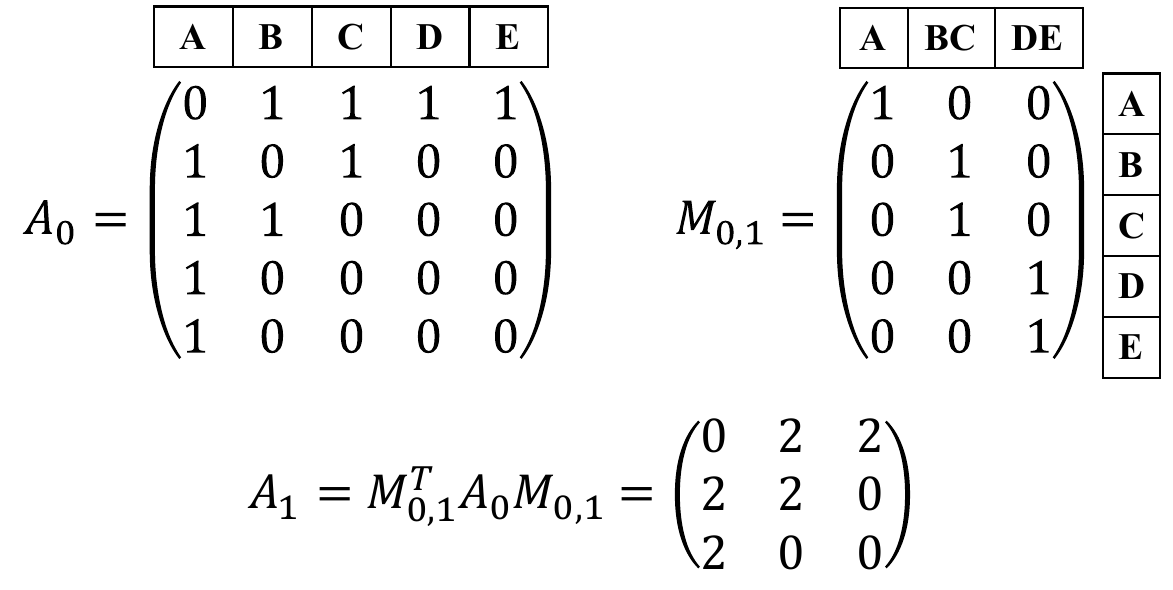}}
%\vspace{-2mm}
\caption[xxx]{\small
Toy example for illustrating graph coarsening\footnotemark.
(a) shows the process of applying Structural Equivalence Matching (SEM) and Normalized Heavy Edge Matching (NHEM) for graph coarsening.
(b) presents the adjacency matrix $A_0$ of the input graph, the matching matrix $M_{0,1}$ corresponding to the SEM and NHEM matchings, and the derivation of the adjacency matrix $A_1$ of the coarsened graph using Eq.~\ref{eq:coarsen}.} 
\label{fig:toy}
\end{figure*}
\fi

\ifext 
To address the aforementioned problem, we propose a scalable 
MultI-Level Embedding (MILE) framework. % in this paper. Our framework is similar to Metis, MLR-MCL, and Graculus~\cite{METIS,MLR-MCL,GRACLUS}, which are popular multi-level graph clustering algorithm. 

Figure~\ref{fig:overview} shows the overview of our MILE framework,
which contains three key phases: graph coarsening, base embedding, and embeddings refining. 

On the whole, we reduce the size of the graph through repeated coarsening and 
run graph embedding on the coarsest graph, after which we perform embeddings refinement to recover
the embeddings on the original graph.

We summarize some important notations in Table~\ref{tab:notation}
and describe our framework in detail below.
\fi

\noindent The MILE framework comprises three phases: graph coarsening, base embedding, and refinement (see Figure~\ref{fig:overview}), described next. 
\begin{comment}
\textcolor{blue}{COMMENTED OUT\\ Table~\ref{tab:notation} includes common notations used in this paper. }
\end{comment}
%\extvspace{-2mm}

\subsection{Graph Coarsening}
%\extvspace{-1mm}
\label{sec:gc}
In this phase, the input graph $\mathcal{G}$ (or $\mathcal{G}_0$) is repeatedly coarsened into a series of smaller graphs 
$\mathcal{G}_1$, $\mathcal{G}_2$, $...$, $\mathcal{G}_m$ such that $|V_0| > |V_1| > ... > |V_m|$.
In order to coarsen a graph from $\mathcal{G}_{i}$ to $\mathcal{G}_{i+1}$, 
multiple nodes in $\mathcal{G}_{i}$ are collapsed to form super-nodes in $\mathcal{G}_{i+1}$,
and the edges incident on a super-node are the union of the edges on the original nodes in $\mathcal{G}_{i}$.
Here the set of nodes forming a super-node is called a \textit{matching}.
We propose a hybrid matching technique containing two
matching strategies that can efficiently coarsen the graph while retaining the global structure. A toy example is shared in Figure \ref{fig:toy}. \looseness=-1

\begin{figure}[t]
        \begin{center}
        % \vspace{1em}
    % \includegraphics[width=0.45\textwidth]{./Figures/toy_example.pdf}
    % \includegraphics[width=0.45\textwidth]{./Figures/toy_example.pdf}
    \subfloat[Using SEM and NHEM for graph coarsening]{\label{fig:toy1}\includegraphics[width=0.90\linewidth]{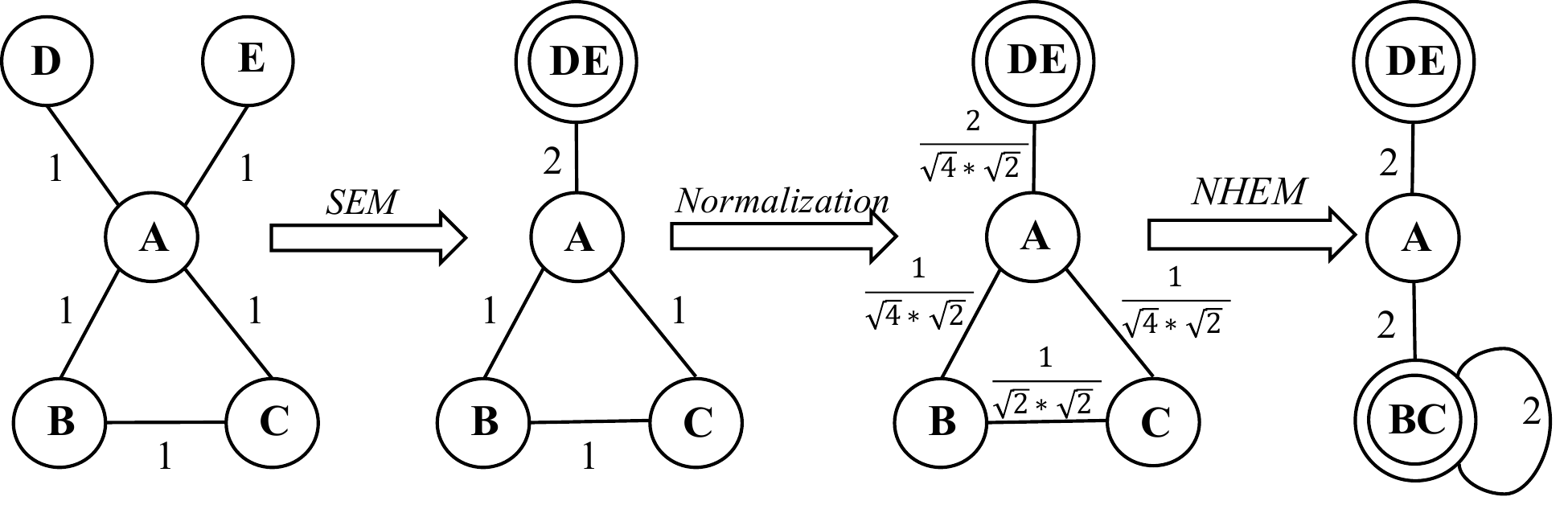}} \hspace{2mm}
    % \vspace{0.5em}
    \subfloat[Adjacency matrix and matching matrix]{\label{fig:toy2}\includegraphics[width=0.8\linewidth]{./Figures/toy2_V.pdf}}
        \end{center}
    \vspace{-1em}    
    \caption[xxx]{
        Toy example for illustrating graph coarsening.
        (a) shows the process of applying Structural Equivalence Matching (SEM) and Normalized Heavy Edge Matching (NHEM) for graph coarsening.
        (b) presents the adjacency matrix $A_0$ of the input graph, the matching matrix $M_{0,1}$ corresponding to the SEM and NHEM matchings, and the derivation of the adjacency matrix $A_1$ of the coarsened graph using Eq.~\ref{eq:coarsen}.
} 
    \label{fig:toy}
\vspace{-1em}
\end{figure}

\ifext
\footnotetext{We follow the strategy in existing work~\cite{MLR-MCL} for weighting self-loops (the weight of the self-loop on node $BC$ is $2$ instead of $1$).}
\fi

%\extvspace{-1mm}

% \noindent

\noindent
\textbf{Structural Equivalence Matching (SEM)} : Given two vertices $u$ and $v$ in an unweighted graph $\mathcal{G}$, we call they are \textit{structurally equivalent} if they are incident on the same set of neighborhoods. In figure \ref{fig:toy1}, node D and E are \textit{structurally equivalent}. The intuition of \textit{matching} structually equivalent nodes is that if two vertices are structurally equivalent, then their node embeddings will be similar. The novel SEM method -- though dependent on graph structure --  is highly effective on the real world datasets.
 Empirically, we found 5-20\% of nodes to be structurally equivalent in many real world datasets. For example, during the first level of coarsening, YouTube has 172,906 nodes (or 86,453 pairs) out of 1,134,890 nodes that are found to be SEM (around 15\%); Yelp has 875,236 nodes (or 437,618 pairs) out of 8,938,630 nodes are SEM (around 10\%). Note that even more nodes are involved in SEM 
across iterations, as SEM is run iteratively at each coarsening level.
%\end{tcolorbox}

\ifext 
\begin{theorem}
% \extvspace{-2mm}
If two vertices $u$ and $v$ in an unweighted graph $\mathcal{G}$ are structurally equivalent,
then their node embeddings derived from $\mathcal{G}$ will be identical.
\label{theorem:equ}
% \extvspace{-2mm}
\end{theorem}

This can be proved by reasoning on the fact that the two nodes are non-distinguishable and interchangeable on $\mathcal{G}$ if they share the same set of the neighborhoods (details of proof omitted due to the limit of space). 
Base on Theorem~\ref{theorem:equ}, we define a \textit{structural equivalence matching} as a set of nodes that are structurally equivalent to each other.
\ifext 
For the example in Figure~\ref{fig:toy1}, nodes $D$ and $E$ are considered a structural equivalent matching.
\fi
\fi
% When constructing a structural equivalence matching, we ignore the weights of the edges.
%\extvspace{-1mm}
\noindent
\textbf{Normalized Heavy Edge Matching (NHEM)} : Heavy edge matching is a popular matching method for graph coarsening~\cite{METIS}. 
We select an unmatched node, say $u$, in the graph and find a large weighted edge  $(u, v)$ incident on node $u$ such that node $v$ is also unmatched. We then collapse nodes $u$ and $v$ into one super-node and mark them as matched. We repeat the matching process until there are no unmatched nodes or an unmatched node does not have an unmatched neighbor node. %  For an unmatched node $u$ in $\mathcal{G}_{i}$, its heavy edge matching is a pair of vertices $(u, v)$ such that the weight of the edge between $u$ and $v$ is the largest. 
In this paper, we propose to normalize the edge weights when applying heavy edge matching using the formula as follows
\begin{equation}
W_i(u, v) = \frac{A_i(u, v)}{\sqrt{D_i(u, u) \cdot D_i(v, v)}}.
\label{eq:norm_wgt}
\end{equation}
\ifext
In Eq.~\ref{eq:norm_wgt},
\else
Here,
\fi
the weight of an edge is normalized by the degree of the two vertices on which the edge is incident.
Intuitively, it penalizes the weights of edges connected with high-degree nodes.
\ifext 
For the example in Figure~\ref{fig:toy1}, 
node $B$ is equally likely to be matched with node $A$ and node $C$ without edge weight normalization.
With normalization, node $B$ will be matched with $C$, which is a better matching since $B$ is more 
structurally similar to $C$. 
\fi
%with normalization, the normalized heavy edge matching of node $B$ is node $C$ instead of node $A$
%for vertex $B$, both vertices $A$ and $C$ are candidates for match. However in terms of structure, vertex $B$ is more similar to %vertex $C$ than vertex $A$.  ??
%As we will show in Sec. ~\ref{sec:embed_refine}, this normalization is tightly connected with the graph convolution kernel.
We will show later that this normalization is tightly connected with the graph convolution kernel.

\noindent
\textbf{Hybrid Matching Method } : We use a hybrid of two matching methods above for graph coarsening. 
To construct $\mathcal{G}_{i+1}$ from $\mathcal{G}_{i}$, 
we first find out all the structural equivalence matching (SEM) $\mathcal{M}_1$, 
where $\mathcal{G}_i$ is treated as an unweighted graph.
This is followed by the searching of the normalized heavy edge matching (NHEM) $\mathcal{M}_2$ on $\mathcal{G}_{i}$.
Nodes in each matching are then collapsed into a super-node in $\mathcal{G}_{i+1}$.
Note that some nodes might not be matched at all and they will be directly copied to $\mathcal{G}_{i+1}$.
\ifext
Figure~\ref{fig:toy1} provides a toy example for illustrating the process.
\fi

%We can then explicitly merge vertices in each matching into a 
%super-node while aggregating edges incident on the original vertices.
Formally, we build the adjacency matrix $A_{i+1}$ of $\mathcal{G}_{i+1}$ through matrix operations.
To this end, we define the \textit{matching matrix} storing the matching information from graph $\mathcal{G}_i$ to $\mathcal{G}_{i+1}$ as a binary matrix $M_{i, i+1} \in \{0, 1\}^{|V_{i}|\times|V_{i+1}|}$. The $r$-th row and $c$-th column
of $M_{i, i+1}$ is set to $1$ if node $r$ in $\mathcal{G}_i$ will be collapsed to super-node $c$ in $\mathcal{G}_{i+1}$, and 
is set to $0$ if otherwise. 
Each column of $M_{i, i+1}$ represents a matching with the $1$s representing the nodes in it.
% entries corresponding to the matching vertices set to $1$ while the remaining are $0$.
Each unmatched vertex appears as an individual column in $M_{i, i+1}$ with merely one entry set to $1$.
\ifext
For the toy example in Figure~\ref{fig:toy}, matching matrix $M_{0,1}$ of dimension $5\times 3$ indicates the mapping information from the original graph to the coarsened graph. In particular, 2nd row and 3rd row means node $B$ and node $C$ form a matching and are mapped to super-node $BC$ in the coarsened graph (similar for 4th and 5th row).
\fi
Following this formulation, we construct the adjacency matrix of $\mathcal{G}_{i+1}$ by using
\begin{equation}
A_{i+1} = M_{i, i+1}^{T}A_{i}M_{i, i+1}.
\label{eq:coarsen}
\end{equation}

Algorithm~\ref{algo:coarsen} summarizes the steps of graph coarsening. 
For each iteration of coarsening,
SEM is generated followed by NHEM (line 2-9). 
A key part of NHEM is to visit the vertices in \textbf{ascending}
order of the number of neighbors (line 5).
This is important to ensure that a large fraction of vertices can be matched in each step, and the graph can be coarsened significantly.
Intuitively, vertices with a small number of neighbors have a limited choice of finding a match and should be given a
higher priority for matching (otherwise, once their neighbors are matched by others, these vertices cannot be matched).
\begin{leftbar}
\end{leftbar}

\begin{comment}
\end{comment}

\noindent {\bf Key Intuitions:}
The graph coarsening phase significantly reduces the size of graph while maintaining the clustering structure and backbone of the original graph.
%amount of memory a method needs to expend thereby offering a clear efficiency gain. 
The coarsening phase also potentially exposes the global structure of the graph to the base embedding method that it otherwise might not take into account thereby offering a potential efficacy gain (up to a point). This efficacy gain has also been observed elsewhere, in the context of stochastic flow algorithms~\cite{MLR-MCL}. Note that stochastic flow algorithms share some common mathematical abstraction to several graph embedding methods ~\cite{DEEPWALK,NODE2VEC,NETMF}. The embeddings learned by base embedding method on the coarsened graph can act as an effective initialization for the graph-topology aware refinement model. %, discussed next.

%\begin{tcolorbox}[breakable, enhanced,colback=yellow!10!white,boxrule=0pt,frame hidden,
%borderline west={1mm}{-2mm}{black}]

\noindent \textbf{Choice for Coarsening Level:}
Similar to other multi-level frameworks, such as graph partitioning~\cite{karypis1998fast,METIS,MLR-MCL} and visualization~\cite{harel2000fast}, the choice of coarsening level depends on the application domain and the graph properties. However, we found a small number of coarsening levels (from $2$ to $4$) usually yields the best-quality embeddings with decent speedup when the graph is of medium size ($\# nodes < 1,000,000$). For larger graph ($\# nodes > 1,000,000$), the embeddings tend to remain constantly high-quality even with coarsening levels from $6$ to $8$ while speedup increases even more. We defer our detailed discussion on this to the experiments section. 
\begin{leftbar}
\end{leftbar}
%\end{tcolorbox}
% Even though graph coarsening techniques have been developed for tasks such as  graph partitioning \cite{karypis1998fast}, graph visualization \cite{harel2000fast} -- the theory supporting graph coarsening is claimed to be circumstantial \cite{loukas2019graph}. 
%The authors \cite{loukas2019graph} quality of different coarsening methods by measuring the closeness of principal eigenvalues of coarsened graph and original graph. 

\begin{algorithm}[!ht]
\small
\caption{Graph Coarsening Algorithm}
\label{algo:coarsen}
\begin{flushleft}
% \vspace{-4mm}
\textbf{Input}: An input graph $\mathcal{G}_0 = (V_0, E_0)$, and coarsening levels $m$. \newline
\textbf{Output}: Coarsened graphs $\mathcal{G}_{i+1}$ and $M_{i, i+1}$ for $0 \le i \le m-1$.
%\vspace{-1mm}
\end{flushleft}
%\vspace{1mm}
\begin{algorithmic}[1]
\For {$i = 1...m$ } % \Comment{node $i$ is labeled.}
  \State $\mathcal{M}_1 \gets$ all the structural equivalence matching in $\mathcal{G}_{i-1}$.
  \State Mark vertices in $\mathcal{M}_1$ as matched.
  \State $\mathcal{M}_2 = \emptyset$. \Comment{storing normalized heavy edge matching}
  \State Sort $V_{i-1}$ by the number of neighbors in ascending order.
  \For {$v \in V_{i-1}$}
    \If {$v$ and $u$ are not matched and $u$ $\in$ Neighbors($v$)}
        \State $(v, u) \gets $ the normalized heavy edge matching for $v$.
        \State $\mathcal{M}_2 = \mathcal{M}_2 \cup (v, u)$, and mark both as matched.
    \EndIf
  \EndFor
  \State Compute matching matrix $M_{i-1, i}$ based on $\mathcal{M}_1$ and $\mathcal{M}_2$. % on $\mathcal{G}_{i}$.
  \State Derive the adjacency matrix $A_{i}$ for $\mathcal{G}_{i}$ using Eq.~\ref{eq:coarsen}.
\EndFor
\State Return $\mathcal{G}_{i+1}$ and $M_{i, i+1}$ for $0 \le i \le m-1$.
\Statex

\end{algorithmic}
%\vspace{-0.4cm}%
\end{algorithm}
% \begin{comment}
% \end{comment}
% \begin{equation}
% D_{i+1} = M_{i}^{T}D_{i}M_{i}
% \end{equation}
\vspace{-1em}

\ifext

\alglanguage{pseudocode}
\ifext

\else 
%\extvspace{-4mm}
\fi

\fi

% \extvspace{-2mm}
\subsection{Base Embedding on Coarsened Graph}
% \extvspace{-1mm}

The size of the graph reduces drastically after each iteration of coarsening, 
halving the size of the graph in the best case. 
We coarsen the graph for $m$ iterations and apply the graph embedding method $f$($\cdot$)
on the coarsest graph $\mathcal{G}_m$. 
%We call this process of embedding the coarsest graph as \textit{base embedding}.
Denoting the embeddings on $\mathcal{G}_m$ as $\mathcal{E}_m$, we have $ \mathcal{E}_m = f ( \mathcal{G}_m\ )
\label{eq:baseEmbed} $. Since our framework is agnostic to the adopted graph embedding method,
we can use any graph embedding algorithm for base embedding.

%, where $m$ is chosen depending on the performance requirement and the size of the original graph. %\footnote{$m$ is usually in the range $[1, 20]$.}.
%At this point, we directly

\begin{figure}
    % \vspace{-2mm}
    \begin{center}
        \includegraphics[width=\linewidth]{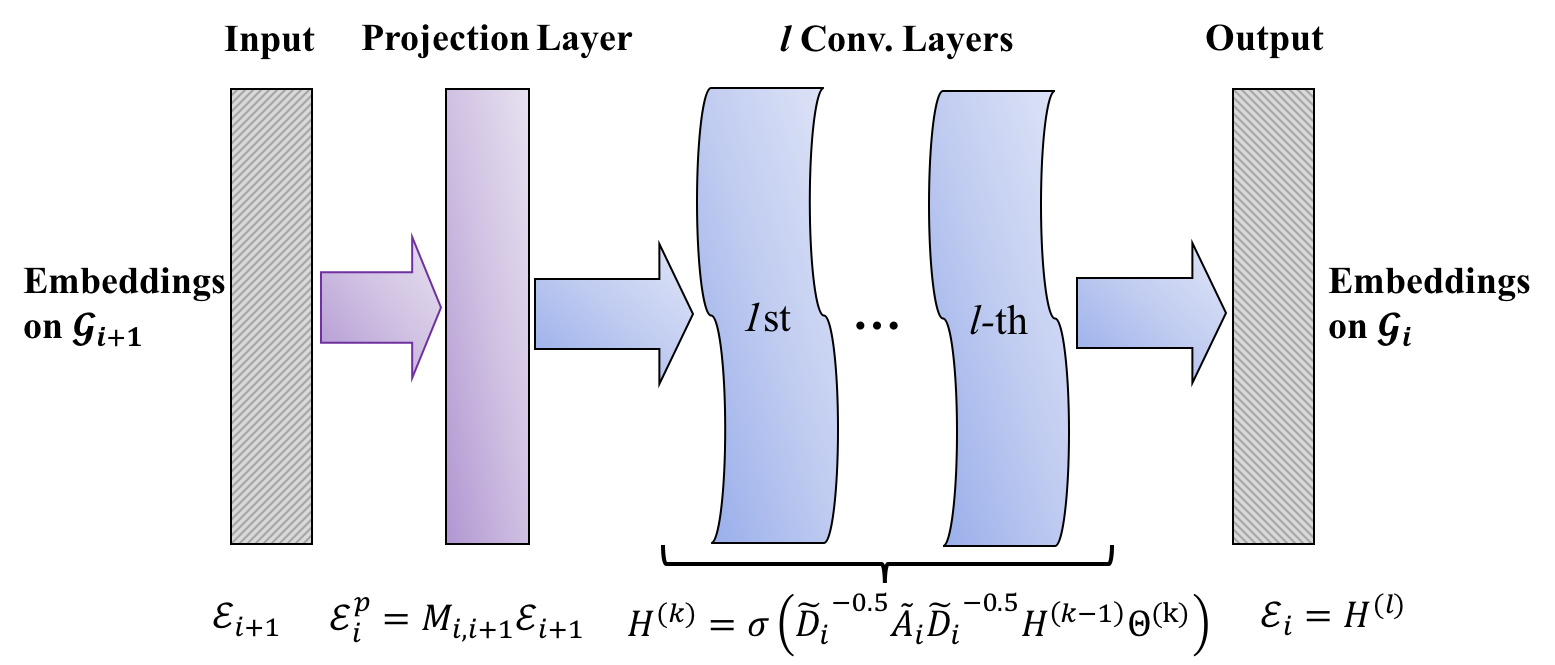}
    \end{center}
%   \vspace{-3mm}
    \caption{Architecture of the embedding refinement model}
    \label{fig:ref_arch}
%    \vspace{-4mm}
\end{figure}

%\extvspace{-2mm}
\subsection{Refinement of Embeddings}
%\extvspace{-1mm}
\label{sec:embed_refine}
The final phase of MILE is the embedding refinement.
Given a series of coarsened graph $\mathcal{G}_0, \mathcal{G}_1, \mathcal{G}_2, ..., \mathcal{G}_m$, their corresponding matching matrix $M_{0,1}, M_{1,2}, ..., M_{m-1, m}$, and the node embeddings $\mathcal{E}_m$ 
on $\mathcal{G}_m$, we seek to develop an approach to derive the node embeddings of $\mathcal{G}_0$ from $\mathcal{G}_m$.
To this end, we first study an easier subtask:
given a graph $\mathcal{G}_i$, its coarsened graph $\mathcal{G}_{i+1}$, the matching matrix $M_{i, i+1}$
and the node embeddings $\mathcal{E}_{i+1}$ on $\mathcal{G}_{i+1}$, how to infer the embeddings $\mathcal{E}_i$
on graph $\mathcal{G}_i$.
Once we solved this subtask,
we can then iteratively apply the technique on each pair of consecutive graphs
from $\mathcal{G}_m$ to $\mathcal{G}_0$ and eventually derive the node embeddings on $\mathcal{G}_0$.
In this work, we propose to use a graph-based neural network model to perform embedding refinement.

%\extvspace{-2mm}
\noindent
\textbf{Graph Convolution Network for  Refinement Learning : }
Since we know the matching information between the two consecutive graphs $\mathcal{G}_{i}$ and $\mathcal{G}_{i+1}$,
we can easily project the node embeddings from the coarse-grained graph $\mathcal{G}_{i+1}$ to
the fine-grained graph $\mathcal{G}_{i}$ using
\begin{equation}
\mathcal{E}^p_{i} = M_{i, i+1} \mathcal{E}_{i+1}
\label{eq:proj}
\end{equation}
In this case, embedding of a super-node is directly copied to its original node(s).
We call $\mathcal{E}^p_{i}$ the \textit{projected embeddings} from $\mathcal{G}_{i+1}$ to $\mathcal{G}_{i}$, or simply \textit{projected embeddings} without ambiguity. 
While this way of simple projection maintains some information of node embeddings,
it has an obvious limitation that 
nodes will share the same embeddings 
if they are matched and collapsed into a super-node during the coarsening phase.
This problem will be more serious when the embedding refinement is performed iteratively from 
$\mathcal{G}_m$, ..., $\mathcal{G}_0$. 
To address this issue, we propose to learn a \textit{graph convolution network} (GCN) for embedding refinement~\cite{GCN}.
Specifically, we design a graph-based neural network model $\mathcal{E}_i = 
\mathcal{R} $($\mathcal{E}^p_{i}, A_i$),  which derives the embeddings $\mathcal{E}_i$ on graph $\mathcal{G}_i$ based on the projected embeddings  $\mathcal{E}^p_{i}$ (from the base method) and the graph adjacency matrix $A_i$ (from the input graph).

\begin{comment}

we consider the graph convolution~\cite{henaff2015deep} of $d$-channel input signals $X$ with filters $g$ on $G$ as
\begin{equation}
X \ast_G g = U \theta_g U^T X.
\label{eq:conv}
\end{equation}
Here, $\theta_g = \text{diag}(\theta)$ is parameterized by spectral multipliers $\theta \in \mathbb{R}^{|V|}$ in the Fourier domain, $U$ is the matrix of eigenvectors of the normalized graph Laplacian $L = I - D^{-\frac{1}{2}} A D^{-\frac{1}{2}}$, where $D$ is a diagonal matrix with entries
$D(i, i) = \sum_j A(i, j)$.
% = U\Lambda U^T$.

Since Eq.~\ref{eq:conv} can be computationally expensive, 
we use its fast approximate version from~\cite{GCN}:
\begin{equation}
% g_{\theta} \ast X 
X \ast_G g \approx \tilde{D}^{-\frac{1}{2}} \tilde{A} \tilde{D}^{-\frac{1}{2}} X  \Theta 
\end{equation}
% \begin{equation}
% g_{\theta} \ast X \approx \theta (I + D^{-\frac{1}{2}} A D^{-\frac{1}{2}}) X
% \end{equation}
% To alleviate the issue of exploding or vanishing gradients, renormalization trick is used so that
where $\tilde{A} = A + \lambda D$, $\tilde{D}(i, i) = \sum_j \tilde{A}(i, j)$, $\Theta \in \mathbb{R}^{d\times d}$, and $\lambda \in [0, 1]$ is a hyper-parameter for controlling the weight of self-loop. 
As this approximate convolution model can be regarded as a layer-wise linear model,
we can stack multiple such layers to achieve a model of higher capacity. 
\end{comment}

Given graph $G$ with adjacency matrix $A$, we consider the fast approximation of graph convolution from \cite{GCN}. The $k$-th layer of this neural network model is
\begin{equation}
H^{(k)}(X, A) = \sigma \left( \tilde{D}^{-\frac{1}{2}} \tilde{A} \tilde{D}^{-\frac{1}{2}} H^{(k-1)}(X, A) \Theta^{(k)} \right)
\label{eq:layer}
% \extvspace{-0.5em}
\end{equation}
where $\sigma(\cdot)$ is an activation function, $\Theta^{(k)}$ is a layer-specific trainable weight matrix, and $H^{(0)}(X, A) = X$. In this paper, we define our embedding refinement model as a $l$-layer graph convolution model
\begin{equation}
\mathcal{E}_i = \mathcal{R} \left(\mathcal{E}^p_{i}, A_i\right) \equiv H^{(l)} \left(\mathcal{E}^p_{i}, A_i \right).
\label{eq:refine}
\end{equation}

The architecture of the refinement model is shown in Figure~\ref{fig:ref_arch}. The intuition behind this refinement model is to integrate the structural information of the current graph $\mathcal{G}_i$
into the projected embedding $\mathcal{E}^p_{i}$ by repeatedly performing the spectral graph convolution. 
%To some extent, 
Each layer of graph convolution network in Eq.~\ref{eq:layer} can be regarded as one iteration of embedding propagation
in the graph following the re-normalized adjacency matrix $\tilde{D}^{-\frac{1}{2}} \tilde{A} \tilde{D}^{-\frac{1}{2}}$.
Note that this re-normalized matrix is well aligned with the way we conduct normalized heavy edge matching in Eq.~\ref{eq:norm_wgt}.

 \noindent \textbf{Choice for Number of GCN Layers:}
The graph convolution model is often treated as a message passing operator \cite{GRAPHSAGE} with the number of layers corresponding to the number of hops in the graph. In other words, $l$ GCN layers correspond to aggregating structural information from all the $l$-hop neighbours for each node. 
On the one hand, we want $l$ to be larger than $1$ so that node embeddings can reflect connectivity structure beyond immediate neighbors. On the other hand, we also do not want too large an $l$ as it will make the node embeddings homogeneous and less distinguishable across the graph due to the small-world property of real-world graphs. Similar to existing literature~\cite{GCN,GRAPHSAGE,velivckovic2017graph} we find that setting $l$ to $2$ worked best in practice.
\begin{leftbar}
\end{leftbar}
%,
%where we apply the same way of re-normalization on the adjacency matrix for edge matching. 
%However, the graph convolution model goes beyond just simple propagation in that
%the activation function is applied for each iteration of propagation and
%each dimension of the embedding interacts with other dimensions controlled by the weight matrix $\Theta^{(k)}$.
% We next discuss how the weight matrix $\Theta^{(k)}$ is learned.

% Therefore, the refining rule is
% \begin{equation}
% \mathcal{E}_{i} = H^{l} \left(\mathcal{E}^p_{i} \right) = H^{l} \left( M_{i} \mathcal{E}_{i+1} \right)
% \end{equation}

\noindent
\textbf{Intricacies of Refinement Learning : }
\label{sec:refine}
The learning of the refinement model is essentially learning
$\Theta^{(k)}$ for each $k \in [1, l]$ according to Eq.~\ref{eq:layer}.
Here we study how to design the learning task and construct the loss function. Since the graph convolution model $H^{(l)}(\cdot)$ aims to predict the embeddings $\mathcal{E}_i$ on graph $\mathcal{G}_i$, we
can directly run a base embedding on $\mathcal{G}_i$ to generate the ``ground-truth'' embeddings
and use the difference between
these embeddings and the predicted ones as the loss function for training. 
We propose to learn $\Theta^{(k)}$ on the coarsest graph and \textbf{reuse} them across all the levels for refinement.
%Specifically, given the coarsest graph $\mathcal{G}_m$, we first perform base embedding to get $\mathcal{E}_m = f(\mathcal{G}_m)$, which serves as the ``ground truth'' for embeddings refinement.
%We then further coarsen graph $\mathcal{G}_m$ into graph $\mathcal{G}_{m+1}$ and perform another base embedding: 
%$\mathcal{E}_{m+1} = f(\mathcal{G}_{m+1})$. 
%Following the embedding refinement procedures, we can predict the refined embeddings on $\mathcal{G}_m$ as
%$\mathcal{R} $($\mathcal{E}^p_{m}, A_m$) $= H^{(l)}(M_{m, m+1} \mathcal{E}_{m+1}, A_m)$.
% We construct a loss function to measure the loss of information when constructing $\mathcal{E}_m$ from $\mathcal{E}^p_{m}$.
%Considering the ``ground truth'' from base embedding and inferred embeddings from the refinement model, 
Specifically, we can define the loss function as the mean square error as follows
\begin{equation}
L = \frac{1}{|V_m|} \left\Vert \mathcal{E}_{m} - H^{(l)}(M_{m, m+1} \mathcal{E}_{m+1}, A_m) \right\Vert^2.
\label{eq:loss1}
% \extvspace{-0.5em}
\end{equation}

We refer to the learning task associated with the above loss function as \textit{double-base} embedding learning.
%since it requires conducting two times of base embedding in the consecutive layers.
We point out, however, there are two key drawbacks to this method.
First of all, the above loss function requires one more level of coarsening to construct $\mathcal{G}_{m+1}$ and an extra base embedding on $\mathcal{G}_{m+1}$.
These two steps, especially the latter, introduce non-negligible overheads to the MILE framework.
More importantly, $\mathcal{E}_{m}$ might not be a desirable ``ground truth'' for the refined embeddings. %, which are predicted based on $\mathcal{E}_{m+1}$.
This is because most of the embedding methods are invariant to an orthogonal transformation of the embeddings, i.e., the embeddings can be rotated by an arbitrary 
orthogonal matrix~\cite{GRAPHSAGE}. In other words, the embedding spaces of graph $\mathcal{G}_m$ and $\mathcal{G}_{m+1}$ can be totally different since the two base embeddings are learned independently. 
Even if we follow the paradigm in~\cite{HARP} and conduct base embedding on $\mathcal{G}_{m}$ using the simple projected embeddings
from $\mathcal{G}_{m+1}$ ($\mathcal{E}^p_{m}$) as initialization, the embedding space does not naturally generalize and can drift during re-training. 
One possible solution is to use an alignment procedure to force the embeddings to be aligned between the two graphs ~\cite{hamilton2016diachronic}. But it could be very expensive.

In this paper, we propose a very simple method to address the above issues. Instead of conducting an additional level of coarsening,
we construct a dummy coarsened graph by simply copying $\mathcal{G}_m$, i.e., $M_{m,m+1} = I$ and $\mathcal{G}_{m+1} = \mathcal{G}_m$. 
By doing this, we not only reduce one iteration of graph coarsening, but also avoid performing base embedding on $\mathcal{G}_{m+1}$ simply because
$\mathcal{E}_{m+1}=\mathcal{E}_{m}$. Moreover, the embeddings of $\mathcal{G}_m$ and $\mathcal{G}_{m+1}$ are guaranteed to be in the same space in this case without any drift. With this strategy, we change the loss function for model learning as follows
\begin{equation}
L = \frac{1}{|V_m|} \left\Vert \mathcal{E}_{m} - H^{(l)}(\mathcal{E}_{m}, A_m) \right\Vert^2.
\label{eq:loss2}
% \extvspace{-0.5em}
\end{equation}

We minimize the difference between the generated embeddings and the embeddings generated from the refinement model (GCN based) so that the learnt refinement model could then be levered to generate embeddings in other coarsening levels.\hl{} With the above loss function, we adopt gradient descent with back-propagation to learn the 
parameters $\Theta^{(k)}$, $k \in [1, l]$. In the subsequent refinement steps, 
we apply the same set of parameters $\Theta^{(k)}$ to infer the refined embeddings. 
We point out that the training of the refinement model is rather efficient as it is done on the coarsest graph.
% which is usually much smaller than the original graph. 
The embedding refinement process involves merely sparse matrix
multiplications using Eq.~\ref{eq:refine} and is relatively affordable compared to conducting embedding on 
the original graph. With these different components, we summarize the whole algorithm of our MILE framework in Algorithm~\ref{algo:whole}. 
%The appendix contains the time complexity of the algorithm in Section \ref{appendix:time}
% loss-0 vs. loss-2.

% loss-0 vs. loss-2.

% \vspace{1em}
\alglanguage{pseudocode}
\begin{algorithm}[t]
    \small
    \caption{Multi-Level Algorithm for Graph Embedding }
    \label{algo:whole}
    \begin{flushleft}
        % \vspace{-4mm}
        \textbf{Input}: An input graph $\mathcal{G}_0 = (V_0, E_0)$, \# coarsening levels $m$, and a base embedding method $f(\cdot)$. \newline
        \textbf{Output}: Graph embeddings $\mathcal{E}_0$ on $\mathcal{G}_0$.
        %\extvspace{-2mm}
    \end{flushleft}
    %\vspace{1mm}
    \begin{algorithmic}[1]
        \State  Coarsen $\mathcal{G}_0$ into $\mathcal{G}_1, \mathcal{G}_2, ..., \mathcal{G}_m$ using proposed hybrid matching method.
        \State Perform base embedding on the coarsest graph $\mathcal{G}_m$ (See Section.~\ref{eq:baseEmbed}).
        \State Learn the weights $\Theta^{(k)}$ using the loss function in Eq.~\ref{eq:loss2}.
        \For {$i = (m-1)...0$ } % \Comment{node $i$ is labeled.}
        \State Compute the projected embeddings $\mathcal{E}^p_{i}$ on $\mathcal{G}_i$.
        \State Use Eq.~\ref{eq:layer} and Eq.~\ref{eq:refine} to compute refined embeddings $\mathcal{E}_i$.
        \EndFor
        \State Return graph embeddings $\mathcal{E}_0$ on $\mathcal{G}_0$.
        \Statex
%       \vspace{-1em}
    \end{algorithmic}
    
\end{algorithm}

%\vspace{2em}

\noindent
\textbf{Discussion on Reusing $\Theta^{(k)}$ Across All Levels:}
 Similar to GCN, $\Theta^{(k)}$  is a matrix of filter parameters and is of size $d*d$ (where $d$ is the embedding dimensionality). Eq. \ref{eq:layer} 
%  $H^{(k)}(X, A) = \sigma \left( \tilde{D}^{-\frac{1}{2}} \tilde{A} \tilde{D}^{-\frac{1}{2}} H^{(k-1)}(X, A) \Theta^{(k)} \right)$
%  in this paper 
 defines how the embeddings are propagated during embedding refinements, parameterized by $\Theta^{(k)}$ . Intuitively,  $\Theta^{(k)}$  defines how different embedding dimensions interact with each other during the embedding propagation. This interaction is dependent on graph structure and base embedding method, which can be learned from the coarsest level.  Ideally, we would like to learn this parameter $\Theta^{(k)}$  on  consecutive levels. But this is not practical since this could be expensive as the graph gets more fine-grained (and defeat our purpose of scaling up graph embedding). This trick of ``sharing'' parameters across different levels is the trade-off between efficiency and effectiveness. To some extent, it is similar to  GCN~\cite{GCN}, where the authors share the same filter parameters $\Theta^{(k)}$  over the whole graph (as opposed to using different $\Theta^{(k)}$  for different nodes; see Eq (6) and (7) in\cite{GCN}). Moreover, we empirically found this works well and is much more efficient. As we will reveal in the Experiments section, the shared $\Theta^{(k)}$ values do much better than alternative $\Theta^{(k)}$ choices during refinement (see Table \ref{tab2}). 
 %shows that if we do not share $\Theta^{(k)}$ values and use random values %for $\Theta^{(k)}$  during refinements, the quality of embedding is much worse (see baseline MILE-untr). 
 \begin{leftbar}
 \end{leftbar}

\noindent
\textbf{Intuition and Rationale of Embedding Refinements:}
Refining the embeddings from a coarsened graph to its fine-grained graph boils down to two steps: a) projecting the node embeddings back to fine-grained graph based on  correspondence matching; and b) propagating the projected embeddings locally, adjusting them based on the fine-grained graph structure. The first step is relatively straightforward through matrix multiplication (Eq.~\ref{eq:proj}). For the second step, we re-purpose the GCN model to propagate the node embeddings. There are mainly two reasons that GCN performs well in this part. First, the propagation rule in GCN is shown to be a first-order approximation of the \textbf{localized spectral filters} on graphs~\cite{GCN,defferrard2016convolutional} and as a result, the propagated embeddings can  capture the local graph structure. Second, GCN contains learnable $\Theta^{(k)}$ and is capable of modeling the interaction between different embedding dimensions in the propagation process, as discussed above.
\begin{leftbar}
\end{leftbar}

%% file: experiments.tex
%In this section, we conduct extensive experiments to gain more insights on the proposed MILE framework. 
% to fully show the advantage of Multi-Level Embedding framework compared to baseline methods.
% We apply MILE on several large-scale datasets with various base embedding methods.
%\begin{comment}

%\end{comment}

% \extvspace{-2mm}
% \subsection{Experimental Configuration}

%\begin{wraptable}{r}{5.5cm}
% % \vspace{2mm}
% % \centering
% % \resizebox{0.9\columnwidth}{!}{
% \begin{tabular}{c|r|r|r}
%   \hline \hline
%   \multicolumn{1}{c|}{\textbf{Dataset}} & \multicolumn{1}{c|}{\textbf{\# Nodes}} & \multicolumn{1}{c|}{\textbf{\# Edges}} & \multicolumn{1}{c}{\textbf{\# Classes}} \\ \hline
%   PPI     & 3,852         & 38,705        & 50    \\ \hline
%   Blog    & 10,312        & 333,983       & 39    \\ \hline
%   Flickr  & 80,513        & 5,899,882     & 195   \\ \hline
%   YouTube & 1,134,890     & 2,987,624     & 47    \\ \hline
%   Yelp    & 8,938,630     & 39,821,123    & 22    \\ \hline \hline
% \end{tabular}
% % }
% % \vspace{-2mm}
% \caption{Dataset Information}
% \label{tab:dataset}
%\vspace{1em} 
%\end{wraptable}

\noindent
\textbf{Datasets:} The datasets used in our study (See Table~\ref{tab:dataset}) have seen prior use for evaluating representation learning methods \cite{DEEPWALK,NODE2VEC,NETMF} and are primarily drawn from popular social media platforms while one is drawn from a well curated bioinformatics dataset.  We preprocess the Yelp dataset following a procedure described in ~\cite{AANE}\footnote{Raw data: \url{https://www.yelp.com/dataset_challenge/dataset}}.

 \begin{table}[t]

  \centering
  \begin{tabular}{c|r|r|r}
    \toprule
    \multicolumn{1}{c|}{\textbf{Dataset}} & \multicolumn{1}{c|}{\textbf{\# Nodes}} & \multicolumn{1}{c|}{\textbf{\# Edges}} & \multicolumn{1}{c}{\textbf{\# Classes}} \\ \midrule
    PPI     & 3,852         & 38,705        & 50    \\ \hline
    Blog    & 10,312        & 333,983       & 39    \\ \hline
    Flickr  & 80,513        & 5,899,882     & 195   \\ \hline
    YouTube & 1,134,890     & 2,987,624     & 47    \\ \hline
    Yelp    & 8,938,630     & 39,821,123    & 22    \\
    \bottomrule
  \end{tabular} 
%   \vspace{1mm}
  \caption{Dataset Information}
  \label{tab:dataset}
%   \vspace{1em}
 \end{table}
 
%  The appendix contains additional details about preprocessing of datasets (Section \ref{appendix:datasets}), details about selected parameters for baselines (Section~\ref{appendix:baseline}),  details about MILE-specific settings (Section~\ref{appendix:MILE-specificSettings}), system specification (Section~\ref{appendix:SystemSpecification}) and details about Evaluation Metrics (Section~\ref{appendix:EvaluationMetrics}).

\smallskip
\noindent
\textbf{Base Embedding Methods:} 
To demonstrate that MILE can work with
different graph embedding methods, we explore several popular methods for graph embedding.
 
\begin{itemize}%[leftmargin=*]

\item \textbf{Random-walk based methods}: We select DeepWalk~\cite{DEEPWALK} and Node2Vec~\cite{NODE2VEC} as baseline methods for this group. We set the length of random walks as $80$, number of walks per node as $10$, and context windows size as $10$. In Node2Vec, we set $p=4.0$ and $q=1.0$ which we found empirically to generate better results across all the datasets.

\item \textbf{Edge reconstruction based method}: We select Line~\cite{LINE} as it is a representative work in this group. The number of edge samples is set to 100 million.

\item  \textbf{Matrix-factorization based methods} : 
GraRep~\cite{GRAREP} and NETMF ~\cite{NETMF} are two popular methods in this group. For GraRep, we set k=4. We varied this parameter but found this value to work well (also suggested by the original authors~\cite{GRAREP}). For NetMF, we set the window size to $10$ and the rank $h$ to $1024$.

\item \textbf{Deep neural network based methods:} We lever SDNE \cite{SDNE} as an exemplar for this group. We set alpha to 0.2 and beta to 10.0. We varied these parameters and found these values to work well -- these parameter values also work well in original paper. 
% TODO: As reported elsewhere

\item \textbf{Distributed embedding methods:}  We compare MILE against Pytorch-biggraph \cite{lerer2019pytorch}. We set the parameter negative batch sizes to 500 and set learning rate to 0.01. We varied these parameters and found these values to work well -- these parameters fall in the suggested values by the authors \cite{lerer2019pytorch}.
\begin{leftbar}
\end{leftbar}

\end{itemize}

By showing the performance gain of using MILE on top of these methods, we want to ensure the contribution of this work is of broad interest to the community.  We also want to reiterate that these methods are quite different in nature. 
\noindent
\textbf{MILE-specific Settings:} 
For all the above base embedding methods, we set the embedding dimensionality $d$ as $128$.
When applying our MILE framework,
we vary the coarsening levels $m$ from $1$ to $8$ whenever possible.
% Discuss more about the selection of levels here.
For the graph convolution network model,
the self-loop weight $\lambda$ is set to $0.05$, the number of hidden layers $l$ is $2$,
and $\tanh(\cdot)$ is used as the activation function,
the learning rate is set to $0.001$ and the number of training epochs is $200$.
The Adam Optimizer is used for model training.

% \vspace{1em}
% \noindent

% \vspace{1em}
\noindent
\textbf{Evaluation Metrics:} 
We evaluate the quality of the embeddings through multi-label node classification~\cite{DEEPWALK,NODE2VEC} and link prediction \cite{gurukar2019network}. %To evaluate the quality of the embeddings, we follow the typical method in existing work to perform multi-label node classification~\cite{DEEPWALK,NODE2VEC} and link prediction \cite{gurukar2019network}.
Specifically, for node classification, we run a $10$-fold cross validation using the embeddings as features and report the average Micro-F1 and average Macro-F1.
For link prediction, we follow the setup described in \cite{gurukar2019network}. Specifically, we evaluate the link prediction performance in terms of AUROC on 5-fold cross validation and report the average AUROC scores. 

\noindent \textbf{Running time:} We  present end-to-end wallclock time  for scalability analysis. For MILE, the reported running time include the execution time of all the phases, including the training time of refinement model.

\noindent
\textbf{System Specifications:} 
The experiments were conducted on a machine running Linux with an
Intel Xeon E5-2680 CPU (28 cores, 2.40GHz) and 128 GB of RAM.
We implement our MILE framework in \texttt{Python}.
% Our code and data are will be available for the replicability purpose.
%with instructions for adding new base embedding methods at \url{http://jiongqianliang.com/MILE/}.
%\footnote{\textbf{Our code and data are available} with instructions for adding new base embedding methods at \url{http://jiongqianliang.com/MILE/}}.
For all the five base embedding methods, we adapt the original code from the
authors
\footnote{DeepWalk: \url{https://github.com/phanein/deepwalk}; \\ Node2Vec: \url{http://snap.stanford.edu/node2vec/}; \\
        Line:  \url{https://github.com/thunlp/OpenNE}
        %\url{https://github.com/tangjianpku/LINE} ; \\ % 
        \\ 
          GraRep: \url{https://github.com/thunlp/OpenNE}; \\ NetMF: \url{https://github.com/xptree/NetMF}}. For SDNE, we lever the publicly available source code from here  \cite{goyal2018graph}.
    We additionally use the TensorFlow package for the embedding refinement learning component.
    We lever the available parallelism (on 28 cores)
    for each method
    (e.g., the generation of random walks in DeepWalk and Node2Vec, the training of the refinement model in MILE, etc.).
% Our code and data are available with instructions for adding new base embedding methods at \url{https://github.com/jiongqian/MILE}.

\subsection{MILE Framework Performance}
\label{sec:exp1}

\begin{figure*}[]
  \centering
  
  \includegraphics[width=0.95\textwidth]{Figures/LinkPrediction/Legend_LP.png}\\
  \setcounter{subfigure}{0}
  \subfloat[PPI (AUROC)]{\label{fig:PPI-mi}\includegraphics[width=0.25\textwidth]{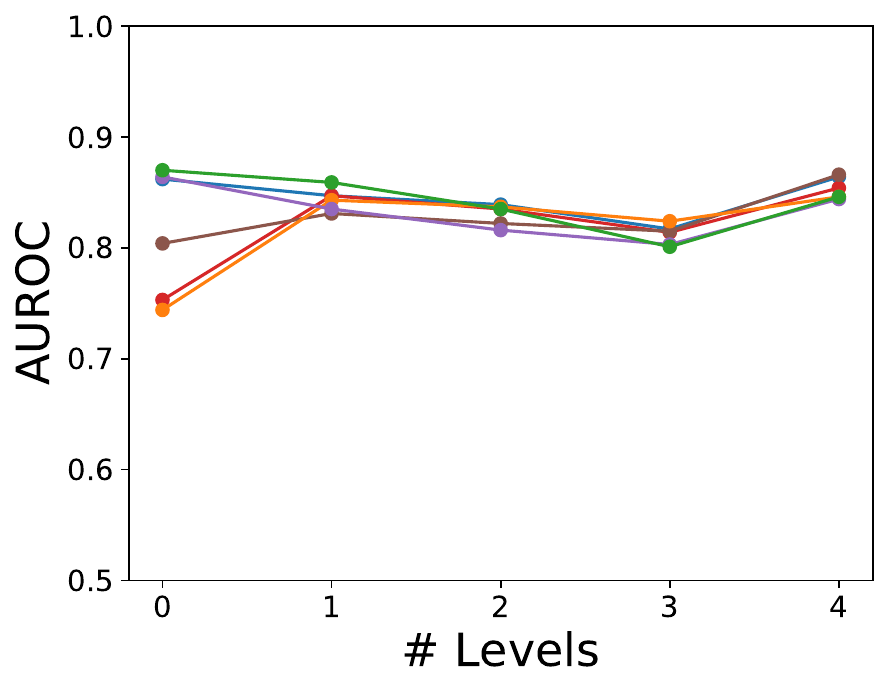}} %\hspace{-1mm}
  \subfloat[Blog (AUROC)]{\label{fig:Blog-mi}\includegraphics[width=0.25\textwidth]{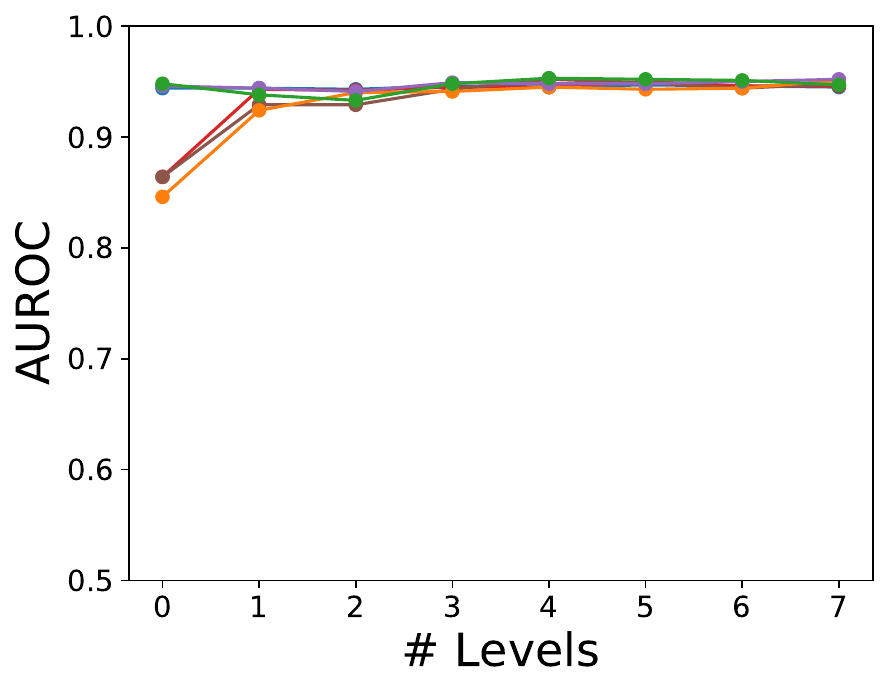}} %\hspace{-1mm}
  \subfloat[Flickr (AUROC)]{\label{fig:Flickr-mi}\includegraphics[width=0.25\textwidth]{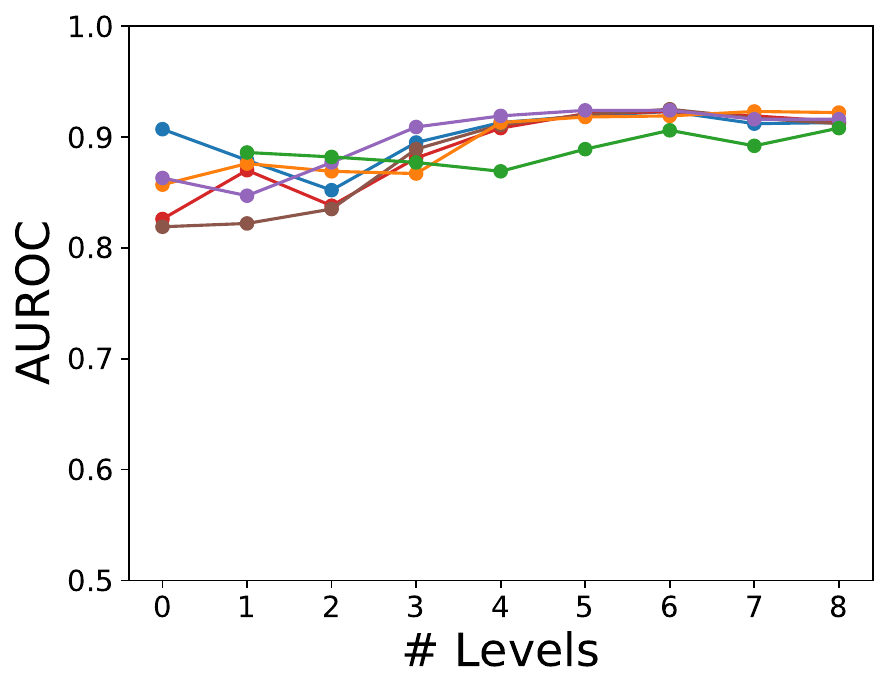}} 
  \subfloat[YouTube (AUROC)]{\label{fig:YouTube-mi}\includegraphics[width=0.25\textwidth]{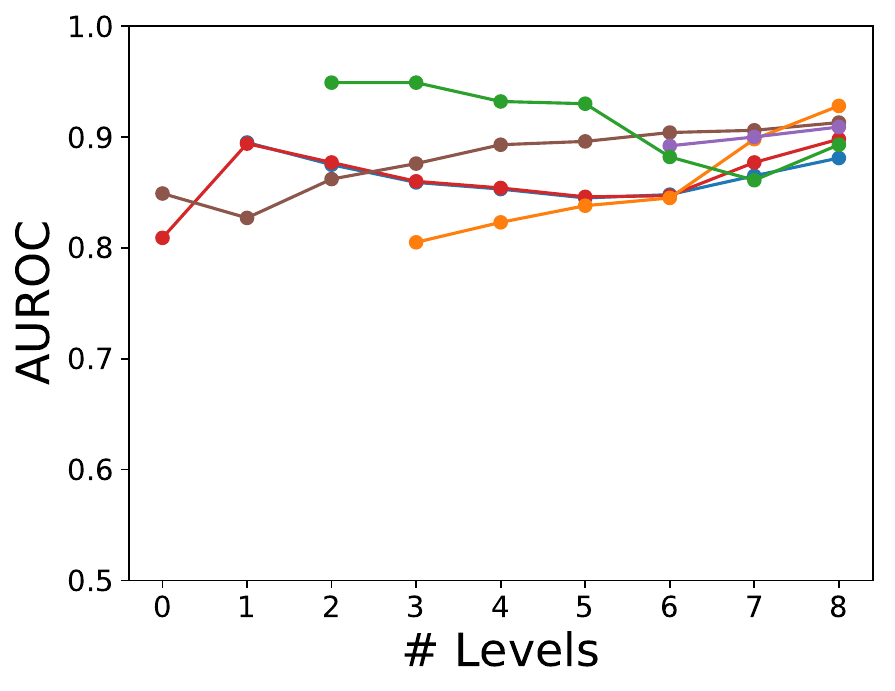}} \\
  \subfloat[PPI (Time)]{\label{fig:PPI-ti}\includegraphics[width=0.25\textwidth]{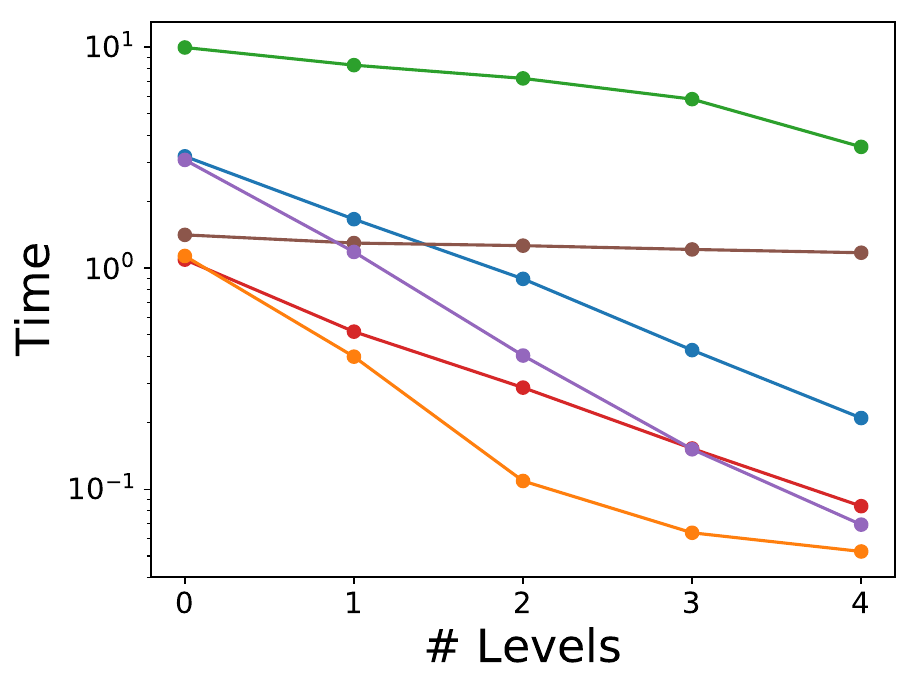}} %\hspace{-1mm}
  \subfloat[Blog (Time)]{\label{fig:Blog-ti}\includegraphics[width=0.25\textwidth]{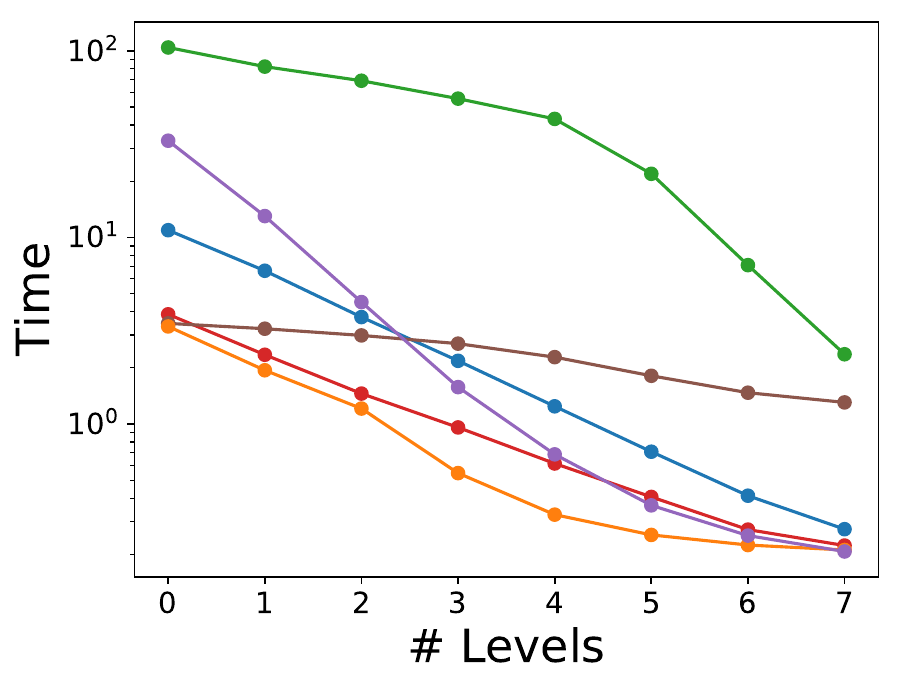}} 
  \subfloat[Flickr (Time)]{\label{fig:Flickr-ti}\includegraphics[width=0.25\textwidth]{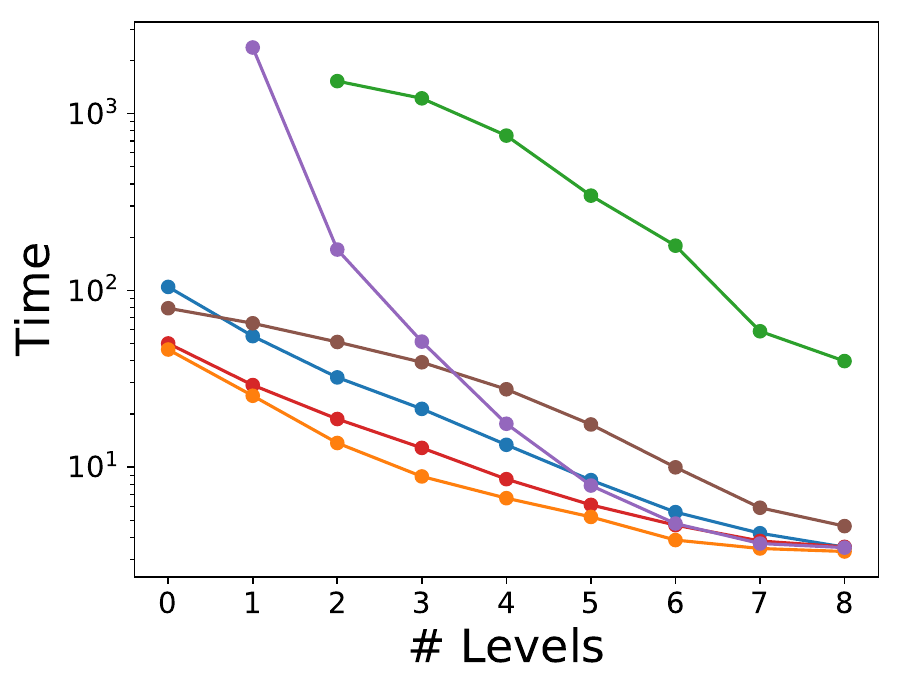}} %\hspace{-1mm}
  \subfloat[YouTube (Time)]{\label{fig:YouTube-ti}\includegraphics[width=0.25\textwidth]{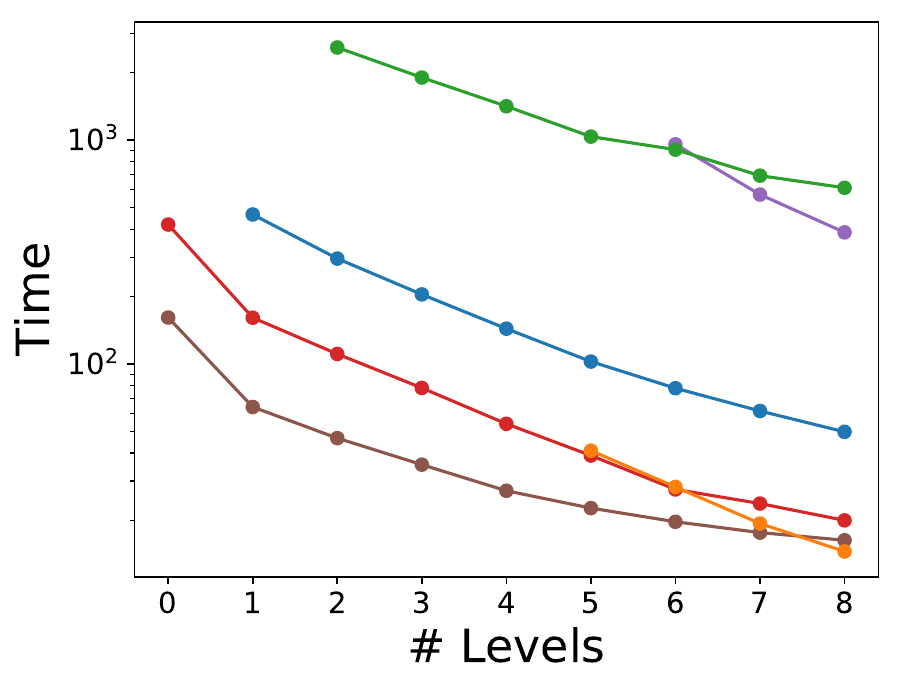}} \\
  
  \caption{
    %\small
     Changes in the link prediction performance as the number of coarsening levels in MILE increases. AUROC and running-time are reported in the first and second row respectively. Running time in minutes is shown in the logarithm scale. Note that \# level $=0$ represents the original embedding method without using MILE. \hl{}
  }
  \label{fig:link_level}
%   \vspace{1em}
  
\end{figure*}

We first evaluate the performance of our MILE framework when applied to different graph embedding methods. %For each dataset, we show the results of MILE under two settings of coarsening levels $m$  and expand on the remaining results in the next section. %Table~\ref{tab1} 
The performance of MILE  with various base embedding methods -- on different datasets and different coarsening levels  -- for node classification and link prediction is shown in Figure~\ref{fig:level} and Figure \ref{fig:link_level}, respectively.\footnote{We discuss the results of Yelp later.} %(we highlight some exact numbers in Table~\ref{tab1}).
We also investigate various design choices related to MILE in a subsequent section. The evaluation of various design choices are conducted through the lens of node classification, but similar results hold for link prediction.
 Note that a coarsening level of $m$=0, corresponds to the original embedding method.  
% In particular, we expand the results of MILE under two settings of coarsening levels $m$ for each dataset in FIGURE. %Table~\ref{tab1}.
%and expand on the remaining results in the next section. 
%Table~\ref{tab1} 
%We select the number of coarsening levels $m$  based on grid search and set $m$ to $1$ and $2$ for PPI, Blog and Flickr, while choosing $6$ and $8$ for YouTube (the interested reader may see Appendix \ref{sec:varied_levels} for details).
We make the following observations:

\begin{figure*}[]
  \centering
  
  \includegraphics[width=0.95\textwidth]{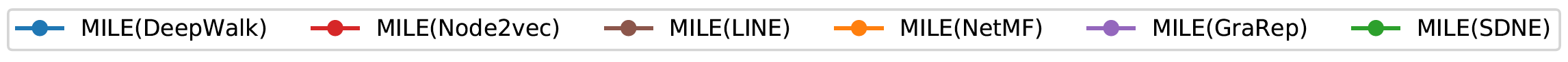}\\
  
  \setcounter{subfigure}{0}
  \subfloat[PPI (Micro-F1)]{\label{fig:PPI-mi}\includegraphics[width=0.25\textwidth]{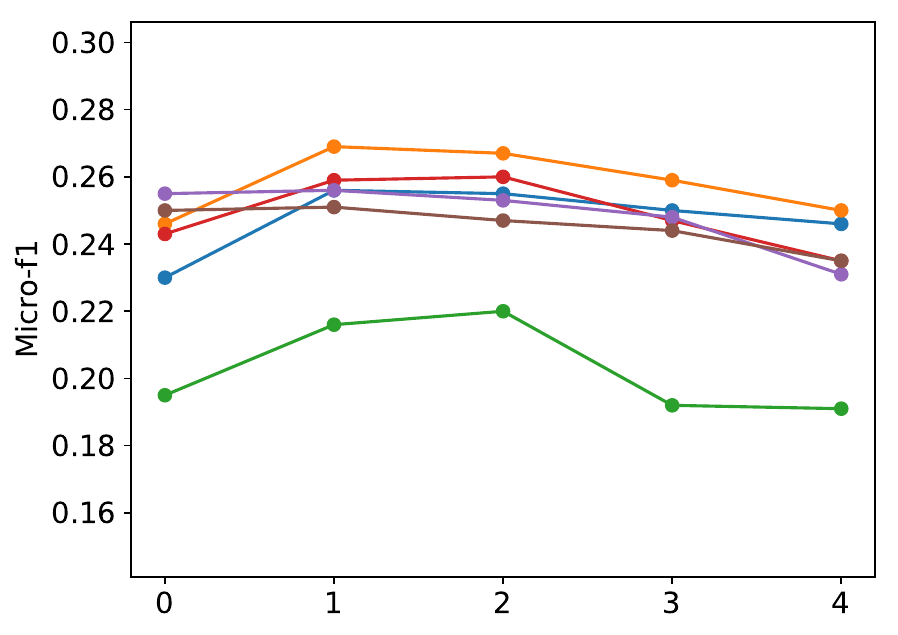}} %\hspace{-1mm}
  \subfloat[Blog (Micro-F1)]{\includegraphics[width=0.25\textwidth]{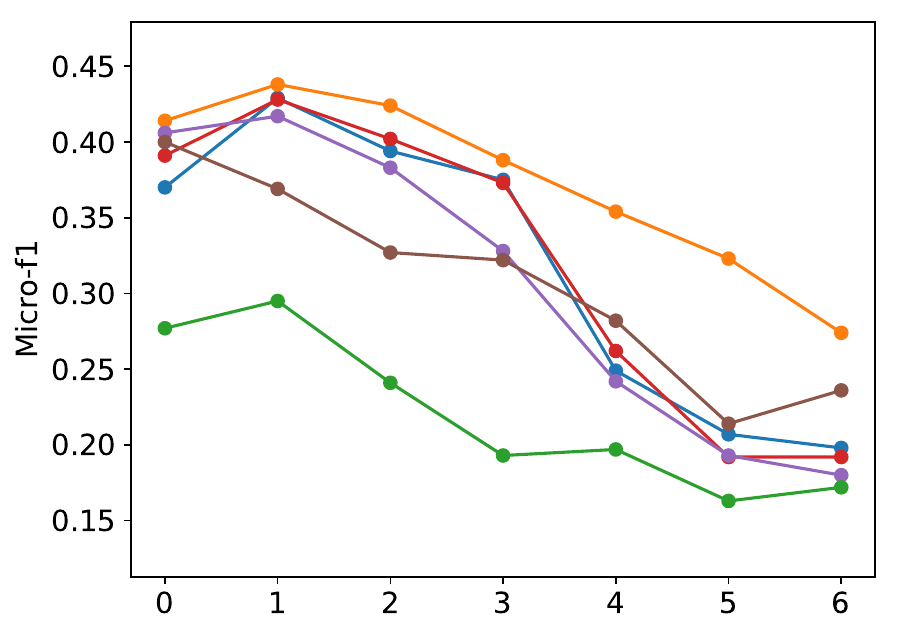}} %\hspace{-1mm}
  \subfloat[Flickr (Micro-F1)]{\label{fig:Flickr-mi}\includegraphics[width=0.25\textwidth]{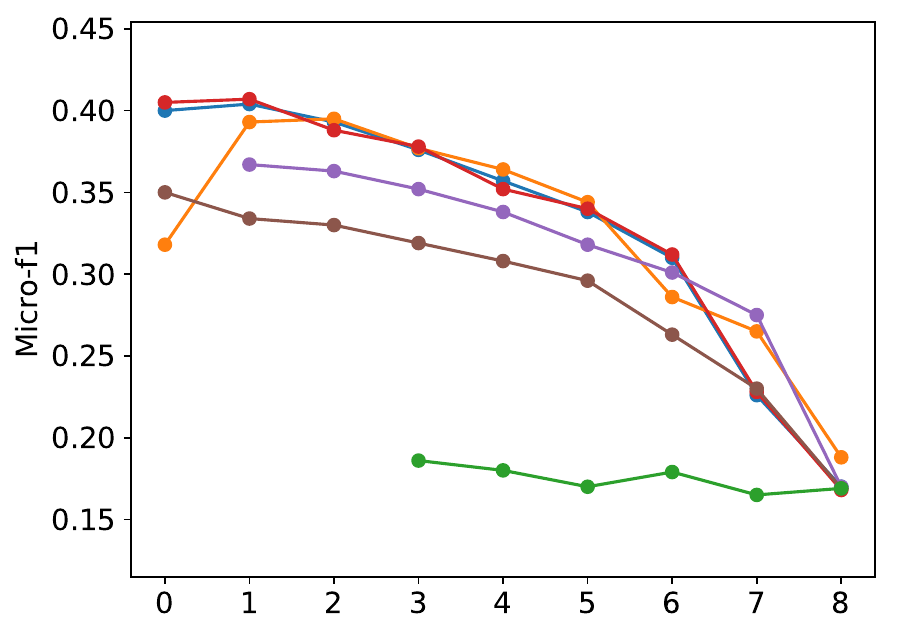}} 
  \subfloat[YouTube (Micro-F1)]{\label{fig:YouTube-mi}\includegraphics[width=0.25\textwidth]{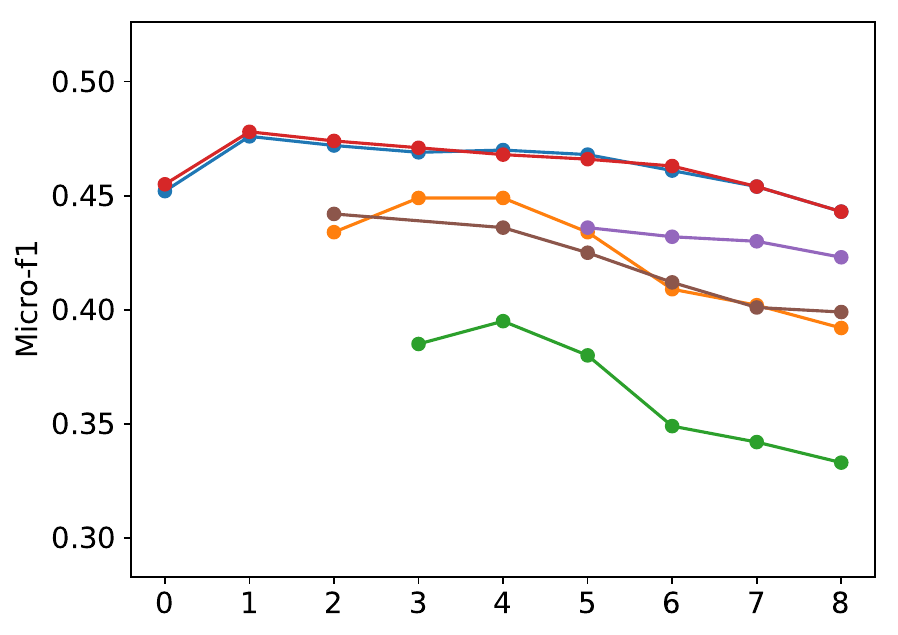}} \\
  \subfloat[PPI (Time)]{\label{fig:PPI-ti}\includegraphics[width=0.25\textwidth]{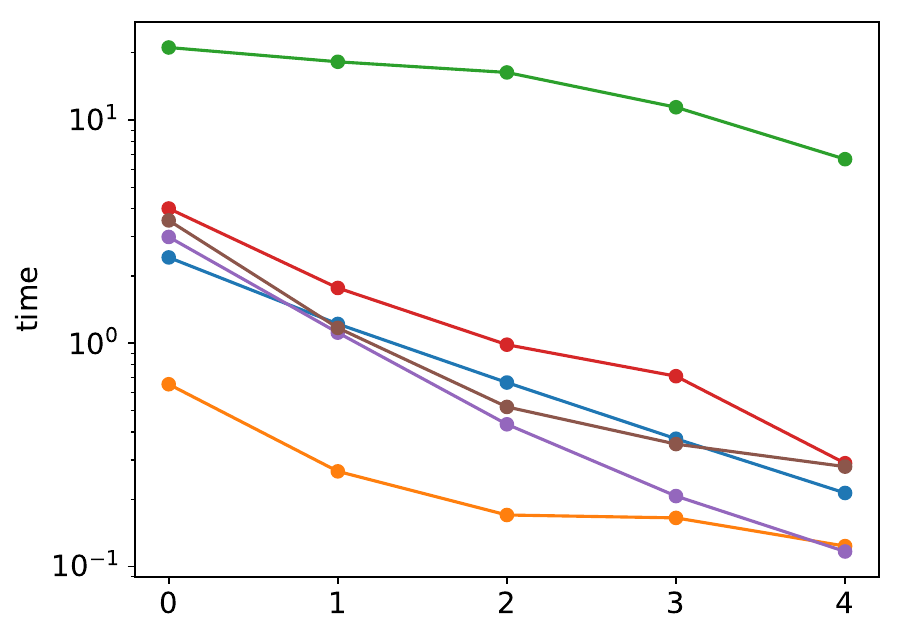}} %\hspace{-1mm}
  \subfloat[Blog (Time)]{\label{fig:Blog-ti}\includegraphics[width=0.25\textwidth]{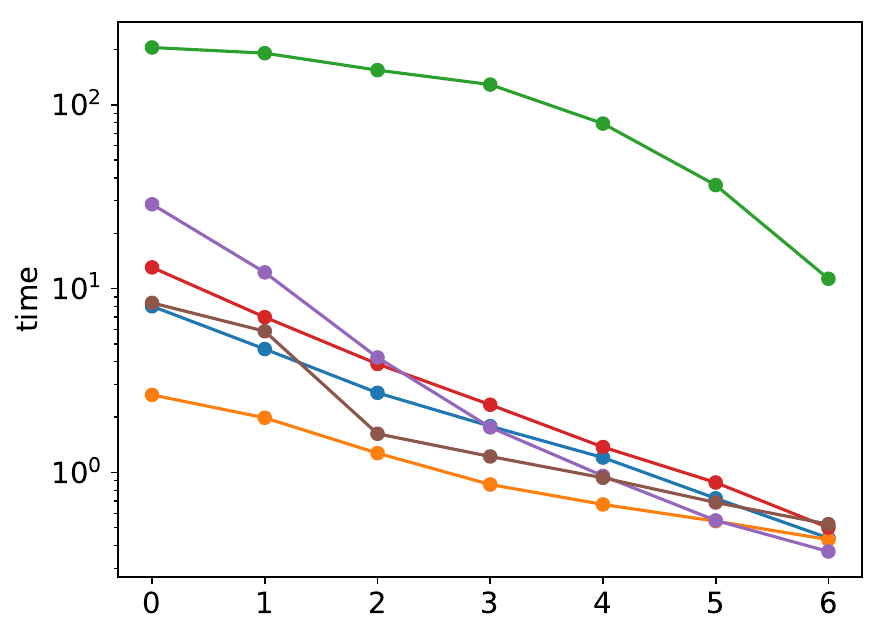}} 
  \subfloat[Flickr (Time)]{\label{fig:Flickr-ti}\includegraphics[width=0.25\textwidth]{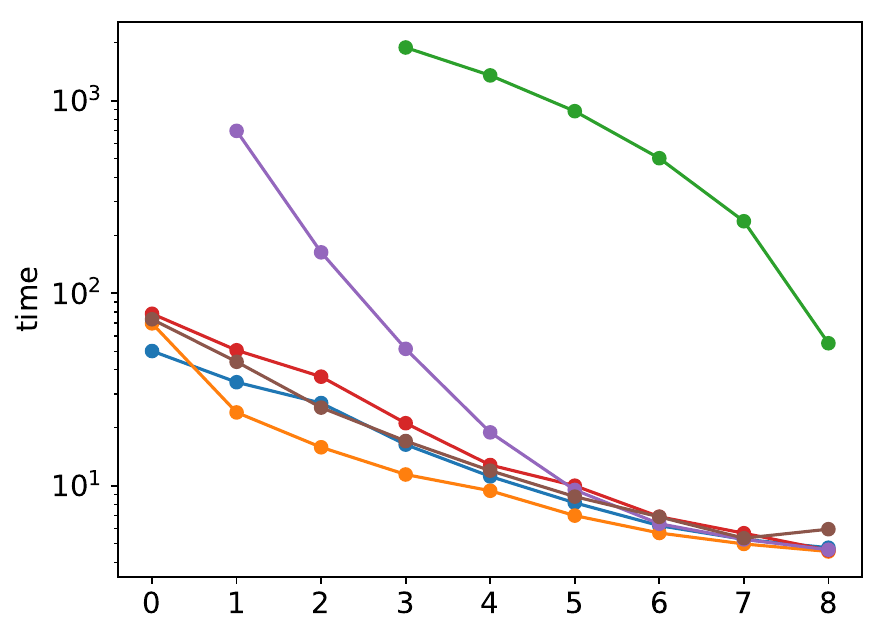}} %\hspace{-1mm}
  \subfloat[YouTube (Time)]{\label{fig:YouTube-ti}\includegraphics[width=0.25\textwidth]{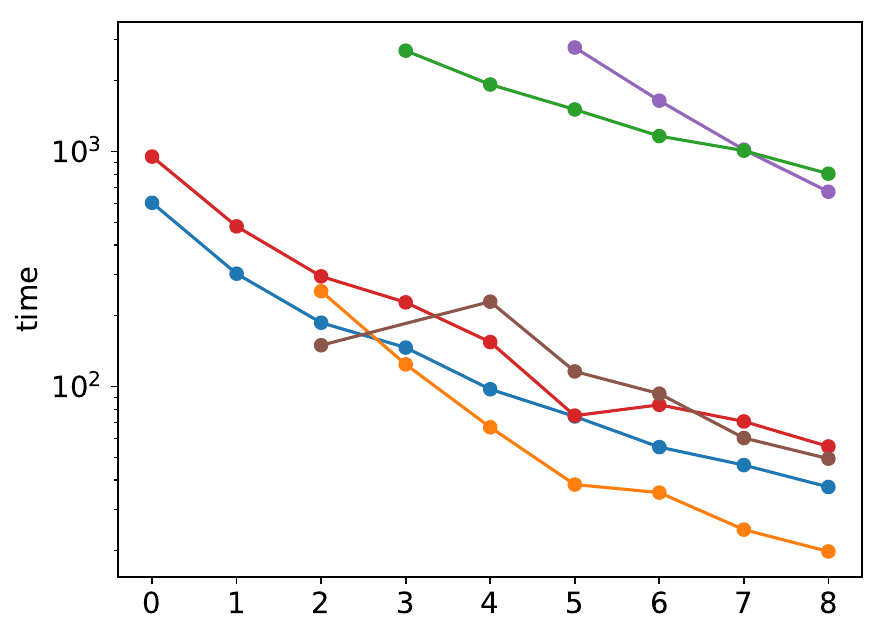}} \\
  
  \caption{
    %\small
    Changes in the node classification performance as the number of coarsening levels in MILE increases (best viewed in color). Micro-F1 and running-time are reported in the first and second row respectively. Running time in minutes is shown in the logarithm scale.  Note that \# level $=0$ represents the original embedding method without using MILE. 
    \hl{}
    Lines/points are missing for algorithms that use over 128 GB of RAM.
  }
  \label{fig:level}
%   \vspace{1em}
  
\end{figure*}

% \begin{itemize}[leftmargin=*]

% \vspace{1em}
% \item \textbf{MILE is scalable}.
\smallskip
\noindent \textbf{MILE is scalable}.
MILE greatly boosts the speed of the explored embedding methods. In the case of node classification, as shown in Figure~\ref{fig:level}, with a single level of coarsening ($m$=1), we are able to achieve speedup ranging from 1.5$\times$ to 3.4$\times$ (on PPI, Blog, and Flickr) while improving qualitative performance. Larger speedups are typically observed on GraRep, NetMF, and SDNE\hl{}. 
Increasing the coarsening level $m$ to $2$, the speedup increases further (up to 14.4$\times$), while the quality of the embeddings is comparable
with the original methods reflected by Micro-F1. % score.
On YouTube, for the coarsening levels $6$ and $8$, we observe more than $10\times$ speedup for DeepWalk, Node2Vec, Line, and SDNE. % {\color{blue}For LINE, we get speedup of at least 1.5$\times$,  while on YouTube dataset, with MILE(LINE, m=8) we get speed up upto 17.46$\times$} 
The execution of SDNE method on flickr dataset does not finish in 2 days, however,  the execution of MILE(SDNE) with coarsen level 8 finishes in less than 1 hour.\hl{}
For NetMF on YouTube, the speedup is even larger -- original NetMF runs out of memory within $9.5$ hours while MILE (NetMF) only takes around  $20$ minutes ($m=8$).  %$35$ minutes ($m=6$) or
In the case of link prediction, as shown in Figure~\ref{fig:link_level}, we observe that an increase in coarsening level results in a consistent decrease in the running time of embedding methods. The higher coarsening levels ($m=$ 7,8) leads to 1.2$\times$ to 113$\times$ speedup for all the methods on BlogCatalog and PPI datasets. On Flickr and YouTube, the running time reduction for all the methods ranges from 10$\times$  to 29$\times$ while the quality of the embeddings is comparable (or even better) with respect to the original methods in terms of AUROC score.

%\footnotetext{The NetMF paper~\cite{NETMF}, reports different results on Flickr with $d=128$ and rank $h=1024$, which we were unable to replicate. In personal communication, its first author promptly acknowledged the error - a much larger rank $h$ is needed to achieve the reported results, which comes at a significant computation and memory cost (their results are on a machine with 1TB of memory).}
% \vspace{1em}
\smallskip
% \noindent \textbf{MILE improves quality}. 
\noindent \textbf{Impact of MILE on embedding quality}. 
In the case of node classification, as shown in Figure~\ref{fig:level}, %for smaller coarsening levels across all the datasets and methods, MILE-enhanced embeddings almost always offer a qualitative improvement over the original embedding method as evaluated by the Micro-F1 score (as high as 24.2\% while many others also show a 10\%$+$ increase). %Evident
for coarsening levels $m=1$ or $2$, we observe that MILE learnt embeddings are almost always better in quality across all the datasets and methods. Examples include MILE (DeepWalk, $m=1$) on Blog/PPI, MILE (Line, $m=1$) on PPI and MILE (NetMF, $m=1$) on PPI/Blog/Flickr. 
%{\color{blue} We see  quality improvements in MILE(LINE) on PPI and YouTube dataset too. However, the speedup of LINE in MILE for Blog and Flickr Dataset also costs in the decrease to quality of embeddings.  }
Even with higher number of coarsening level ($m=2$ for PPI/Blog/Flickr; $m=6,8$ for YouTube),
MILE in addition to being much faster can still improve, qualitatively, 
over the original 
methods on most of the datasets, e.g., MILE (NetMF, $m=2$) $\gg$ NetMF on PPI, Blog, and Flickr. 
In the case of link prediction, as shown in Figure~\ref{fig:link_level}, we observe that the increase in coarsening level results in a corresponding increase in AUROC scores in most of the cases. For Node2vec, LINE and NetMF, with a single level of coarsening ($m$=1), we see improvement in AUROC from 3\% to 10\% on PPI and BlogCatalog datasets. With higher coarsening levels ($m$=7, 8), we see a consistent improvement in AUROC from 1.4\% to 10.7\% for Node2vec, LINE, NetMF and GraRep methods on Flickr and YouTube datasets. For SDNE, on Blog and YouTube dataset, the link prediction performance across coarsening levels 2-5 remains competitive with original SDNE method.\hl{} We observe a 2x-15x speed in this range, consistent with other methods that use MILE. 
As discussed when describing the key intuitions underpinning the coarsening-refinement strategy that MILE employs (in Section~\ref{sec:gc}),
the observed improvement on quality -- for both node classification and link prediction -- is likely due to the fact that the base embedding methods are exposed to a  holistic view of the entire graph. This observation is consistent with those observed for stochastic flow clustering \cite{MLR-MCL}. 
%\input{table1.tex}
%\input{table1short.tex}

% Two m values results go here:
%\input{table1.tex}
% \vspace{1em}
\smallskip
\noindent \textbf{MILE supports multiple embedding methods.} 
We empirically show that MILE can work with different category of embedding methods on multiple real world datasets.
We observe that MILE 
%consistently 
often
improves both the quality and the 
efficiency of NetMF on all four datasets (for YouTube, NetMF 
runs out of memory). For the largest dataset, the speedups afforded exceed 30-fold.
We observe that for GraRep, while speedups with MILE are consistently observed, the qualitative improvements  if any, are smaller (for both YouTube and Flickr, the base method runs out of memory).
For Line, even though its time complexity is linear to the number of edges~\cite{LINE}, 
applying MILE framework on top of it still generates significant speed-up (likely due to the fact that the complexity of Line contains a larger constant factor $k$ than MILE). 
On the other hand, MILE on top of Line generates better quality of embeddings on PPI and YouTube while falling a bit short on Blog and Flickr.   
For DeepWalk and Node2Vec, we again observe consistent improvements in scalability (up to 11-fold on the larger datasets) on the node classification task using MILE with a few levels of coarsening. However, when the coarsening level is increased, the additional
speedup afforded (up to 17-fold) comes at a mixed cost to quality (micro-F1 drops slightly). We observe similar trends -- improvements in embedding quality and reduction in the running time -- in the case of link prediction experiments.

\smallskip
\noindent \textbf{Impact of varying coarsening levels on MILE.} 
%Next, we report our observations as we vary the number of coarsening levels. 
In the case of node classification, as shown in Figure~\ref{fig:level}, when coarsening level $m$ is small ($m=1$ or $2$), MILE tends to significantly improve the quality of embeddings while taking much less time.  From $m=0$ to $m=1$, we see a clear jump of the Micro-F1 score on all the datasets across the  base embedding methods. This observation is more evident on larger datasets (Flickr and YouTube). On YouTube,  MILE (DeepWalk) with $m$=1 increases the Micro-F1 score by $5.3\%$ while only consuming half of the time compared to the original DeepWalk. MILE (DeepWalk) continues to generate embeddings of better quality than DeepWalk until $m=7$, where the speedup is 13$\times$.  As the coarsening level $m$ in MILE increases, the running time drops dramatically while the quality of embeddings only decreases slightly. 
In the case of link prediction, as shown in Figure~\ref{fig:link_level}, the methods with higher coarsening levels ($m$=7,8) on all the evaluated datasets, except GraRep on PPI, show an improvement of AUROC with respect to the embedding performance on the original graph. For SDNE, we observe that as we vary the coarsening levels till coarsen level 5, the link prediction performance on Blog, Flickr, YouTube remains close to original method while we see improvement in the speedup. However on  YouTube we see a drop in AUROC score with coarsening levels 6 and 8 for SDNE. \hl{} For  both node classification and link prediction, the execution time decreases at an almost exponential rate (logarithm scale on the y-axis in the second row of Figure~\ref{fig:level} and Figure~\ref{fig:link_level}). On the other hand, the Micro-F1 score descends much more slowly (the first row of Figure~\ref{fig:level}), most of which are still better than the original methods.

Overall, the above experiments shows that MILE can not only accommodate existing embedding methods - treating them as a blackbox - but also provides nice trade-off between effectiveness and efficency, a useful lever for downstream tasks and use-cases.
%Sacrificing a tiny fraction of quality gain on embeddings (could still be better than original method) can save a huge amount of computational resource.

% \end{itemize}

\begin{comment}
To summarize, our MILE framework not only significantly speeds up the embedding methods, 
but also improves the quality of the node embeddings.
We do notice that it negatively affects the embeddings in some case when the number of coarsening 
levels is large, e.g., MILE (GR, $m=2$) on PPI and Blog. But the decrease is mostly minor compared to 
large speedup achieved. Moreover, we can reduce the coarsening levels in order to generate better embeddings (e.g., $m=1$)
if the quality of the embeddings is valued over efficiency.
We discuss the trade-off between quality and efficiency of graph embedding in Sec.~\ref{sec:varied_levels}
\end{comment}

% \extvspace{-2em}

\subsection{MILE: Large Graph Embedding} %: Embedding a Graph of 40 Million Edges}
\begin{figure}[t]
  \centering
%   \vspace{1em}
  \includegraphics[width=\linewidth]{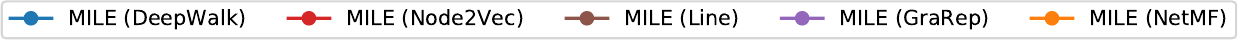}\\
  %\vspace{-3mm}
  \setcounter{subfigure}{0}
  \subfloat[Micro-F1]{\label{fig:yelp-micro}\includegraphics[width=0.48\linewidth]{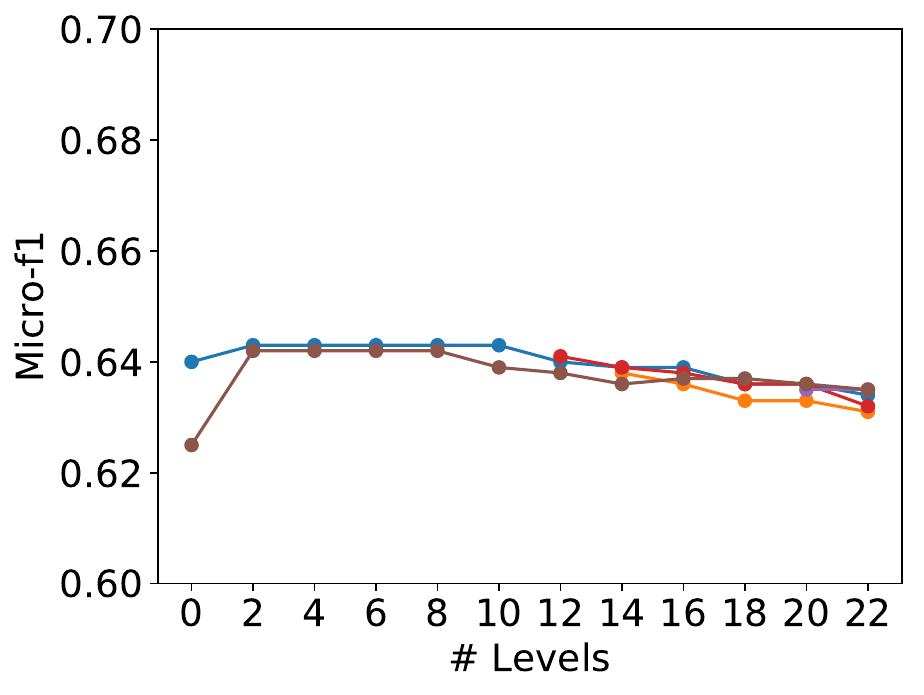}} \hspace{1mm}
  \subfloat[Running Time]{\label{fig:yelp-time}\includegraphics[width=0.48\linewidth]{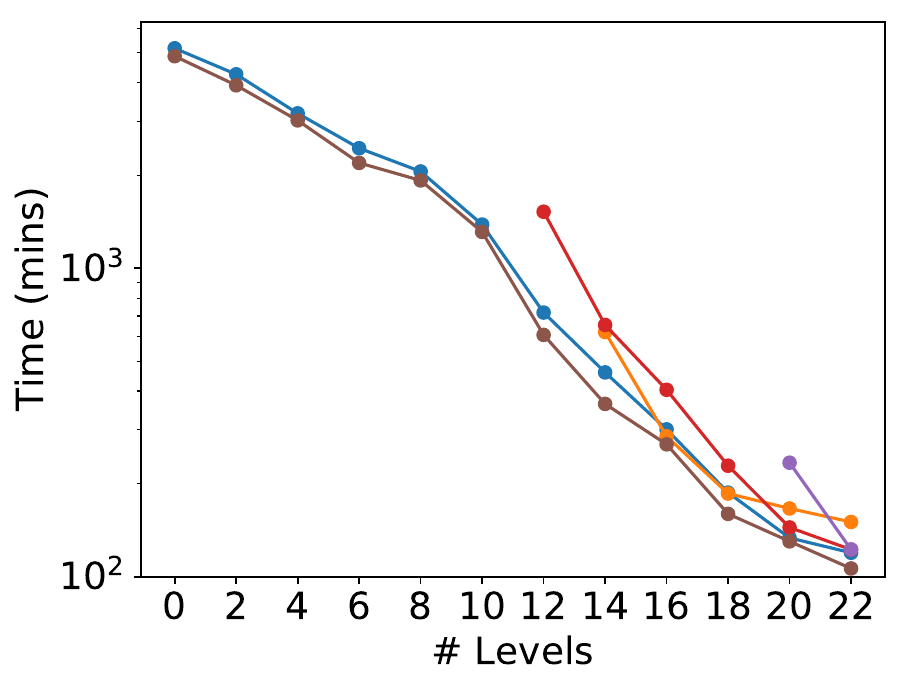}} %\hspace{-1mm}
  %\vspace{-4mm}
  \caption{Running MILE on Yelp dataset. Lines/points are missing for algorithms that do not finish within 60 hours or 
    use over 128 GB of RAM.}
  \label{fig:yelp}
%   \vspace{1em}
\end{figure}

%\vspace{-1em}
We now explore the scalability of MILE on the large Yelp dataset. 
None of the five graph embedding methods studied in this paper can successfully conduct graph embedding on 
Yelp within 60 hours on a modern machine with 28 cores and 128 GB RAM.
Even extending the run-time deadline to 100 hours, we see DeepWalk and Line barely finish. 
%In order to see the scalability and accuracy benefits of MILE, we report the results for DeepWalk and Line on a machine with 32 cores and 1 TB RAM.
Leveraging the proposed MILE framework now makes it much easier to perform graph embedding on this scale of datasets
%To this end, we run the MILE framework on Yelp using the five graph embedding techniques as the base embedding methods with various 
%coarsening levels 
(see Figure~\ref{fig:yelp} for the results).
%from $0$ to $22$. Figure~\ref{fig:yelp} shows the performance of MILE with different base embedding methods on Yelp.
We observe that MILE significantly reduces the running time and improves the Micro-F1 score. 
For example, 
The Micro-f1 scores of original DeepWalk and Line are $0.640$ and $0.625$ respectively, which all take more than 80 hours. But using MILE with $m=4$, the micro-F1 score improves to 0.643 (DeepWalk) and 0.642 (Line) while achieving speedups of around $1.6\times$. 
Moreover, MILE reduces the running time of DeepWalk from $53$ hours (coarsening level $4$) to 
$2$ hours (coarsening level $22$) while reducing the Micro-F1 score just by $1\%$ (from 0.643 to 0.634).
Meanwhile, there is no change in the Micro-F1 score from the coarsening level $4$ to $10$,
where the running time is improved by a factor of two. 
% MILE (DeepWalk) with coarsening level $4$ ({\bf please check}), not shown in table, takes $3,191$ minutes (or 53 hours) with the Micro-F1 score  equal to $64.3$.
These results affirm the power of the proposed MILE framework on scaling up graph embedding algorithms while generating quality embeddings. We also observe similar performance gain and reduction in the running time for the link prediction task, however, due to paucity of space the link prediction results on Yelp dataset are not shared in this paper.

% {\bf: SP : Anything on LINK PREDICTION HERE?}
% Saket: No.

%The appendix contains additional experimental details about MILE for Design choices of coarsening and refinement (Sec.\ref{appendix:design}), effect of varying coarsening levels on accuracy and running time\ref{sec:varied_levels} and memory consumption of embedding techniques with and without MILE \ref{appendix:memory}.

% Figure~\ref{fig:yelp} shows the performance of MILE (DW) with varied coarsening levels on Yelp.
%It is easy to see that the MILE framework significantly reduces the running time from $53$ hours (coarsening level $4$) to 
% $2$ hours (coarsening level $22$) while just reducing the Micro-F1 score by just $1\%$ (from 0.643 to 0.634).

% Meanwhile, there is almost no change in the Micro-F1 score from coarsening level $4$ to $10$,
% where the running time reduce by more than $50\%$. 

\subsection{Memory Consumption}
\label{appendix:memory}
%\extvspace{-1em}

\begin{figure}[t]
	\centering
	\vspace{-2em}
	\setcounter{subfigure}{0}
	\subfloat[MILE (GraRep)]{\label{fig:mem-gr}\includegraphics[width=0.48\linewidth]{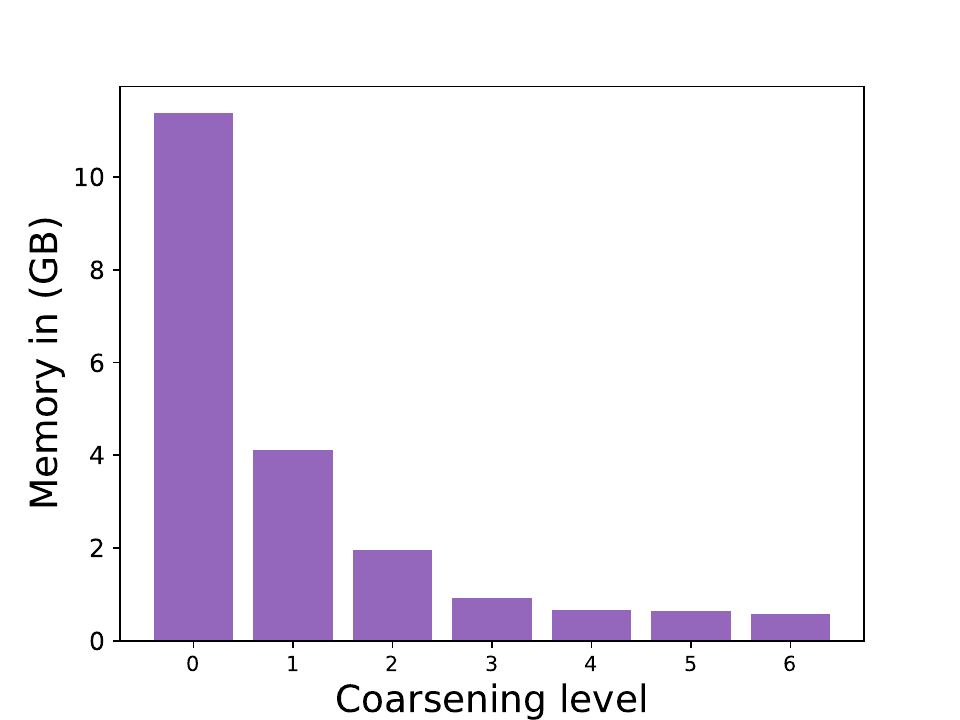}} %\hspace{4mm}
	\subfloat[MILE (NetMF)]{\label{fig:mem-nm}\includegraphics[width=0.48\linewidth]{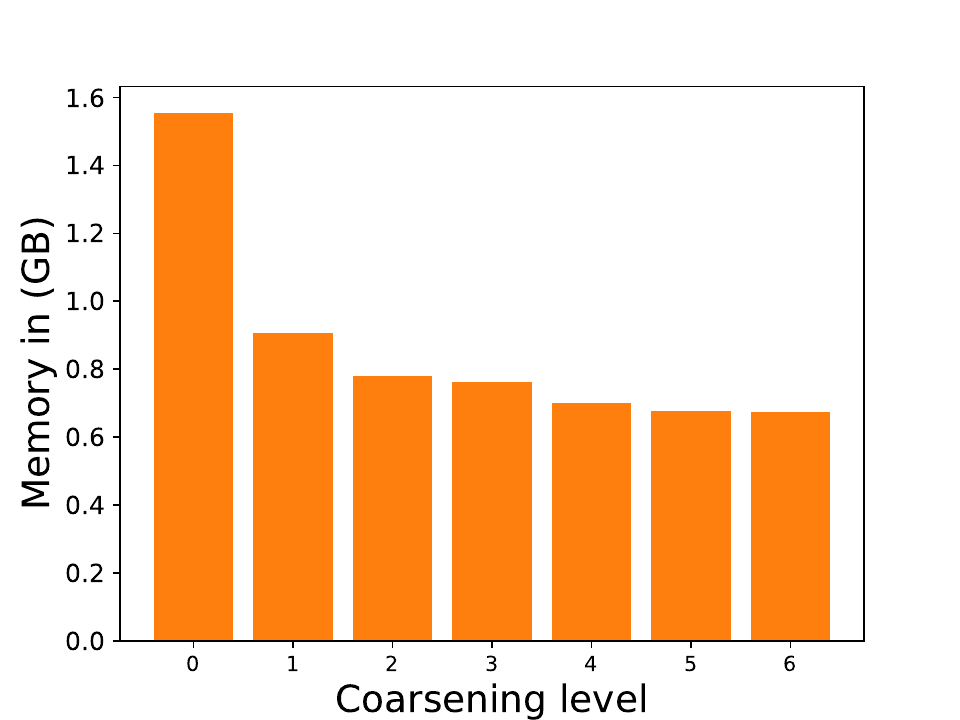}} %\hspace{-1mm}
	% \vspace{-3mm}
	\caption{
		%\small
		Memory consumption of MILE (GraRep) and MILE (NetMF) on Blog with varied coarsening levels. 
		\ifext
		Coarsening level 0 corresponds to the original embedding method without applying the MILE framework.
		\fi
	}
	\label{fig:memory}
	\vspace{-1em}
\end{figure}
We also study the impact of MILE on reducing memory consumption.
For this purpose, we focus on MILE (GraRep) and MILE (NetMF), with GraRep and NetMF as base embedding methods respectively. %MILE (GraRep)
Both of these are embedding methods based on matrix factorization, which possibly involves a dense objective matrix and could be rather memory expensive. We do not explore DeepWalk and Node2Vec here since their embedding learning methods generate truncated random walks (training data) on the fly
with almost negligible memory consumption (compared to the space storing the graph and the embeddings).
% General speaking, the memory consumption of an embedding method can be divided into two parts. 
Figure~\ref{fig:memory} shows the memory consumption of MILE (GraRep) and MILE(NetMF) as the coarsening level increases on Blog (results on other datasets are similar). 
We observe that MILE significantly reduces the memory consumption as the coarsening level increases. 
Even with one level of coarsening, the memory consumption of GraRep and NetMF reduces by $64\%$ and $42\%$ respectively. 
The dramatic reduction continues as the coarsening level increases until it reaches $4$, where the
memory consumption is mainly contributed by the storage of the graph and the embeddings.
This memory reduction is consistent with our intuition, since both \# rows and \# columns in the objective matrix 
\ifext 
for factorization 
\fi
reduce almost by half with one level of coarsening.

\subsection{Comparing MILE with HARP}

%As mentioned in Sec.~\ref{sec:related},
HARP is a  
multi-level method primarily
for improving the quality of graph embeddings.
We compare HARP with our MILE framework using DeepWalk and Node2vec as the base embedding
\ifext
methods\footnote{We use the source code from the authors: \url{https://github.com/GTmac/HARP}. 
  Results on Node2Vec are similar and hence omitted.}.
\else
methods\footnote{\url{https://github.com/GTmac/HARP}}.
\fi
Table~\ref{tab:HARP} shows the performance of these two methods on the four datasets (coarsening level is $1$ on PPI/Blog/Flickr and $6$ on YouTube). 
We also observe similar node classification performance with Macro-f1 metric (not shown).
From the table, we can observe that MILE generates embeddings of comparable quality with HARP.
MILE performs much better than HARP on PPI and Blog, marginally better on Flickr and marginally worse on YouTube.
However, MILE is significantly faster than HARP on all the four datasets (e.g. on YouTube, MILE affords a  31$\times$ speedup).
This is because HARP requires running the whole embedding algorithm on each coarsened graph, which introduces a \textbf{huge computational overhead}. %Note that for PPI and BLOG -- MILE with NetMF (not shown) as its base embeddings produces the best micro-F1 of 26.9 and 43.8, respectively. 
%This shows another advantage of MILE - agnostic to the base embedding when compared with HARP. %see Sec.~\ref{sec:related} for more discussions).

% . Different from MILE, it mainly
% focuses on improving the quality of embeddings by using a better initialization while introducing a non-negligible overhead. 
% \textbf{HARP}~\cite{HARP}\footnote{\url{https://github.com/GTmac/HARP}}. \textbf{working on it...}

\begin{table}[]
  \centering
% \extvspace{-2mm}
    \resizebox{0.85\linewidth}{!}{
      
      \begin{tabular}{c|c|c|c|c}
        \hline
        \hline
        & \multicolumn{2}{c|}{PPI}    & \multicolumn{2}{c}{Blog}    \\
        \hline
        & \multicolumn{1}{c|}{Mi-F1}       & Time      & Mi-F1       & Time       \\
        \hline
        DeepWalk (DW)        & 23.0           & 2.4       & 37.0           & 8.0        \\
        MILE (DW) & 25.6           & 1.2       & 42.9           & 4.6        \\
        HARP (DW) & 24.1           & 3.0       & 41.3           & 9.8        \\
        Node2Vec (NV) & 24.3 & 4.0 & 39.1 & 13.0 \\
        MILE (NV) & {\bf 25.9}           & 1.7       & {\bf 42.8   }        & 6.9        \\
        HARP (NV) & 22.3           & 3.9      & 36.2           & 13.16        \\
        \hline
        \hline
        & \multicolumn{2}{c|}{Flickr} & \multicolumn{2}{c}{YouTube} \\
        \hline
        & Mi-F1       & Time      & Mi-F1       & Time       \\
        \hline
        DeepWalk          & 40.0           & 50.0      & 45.2           & 604.8      \\
        MILE (DW) & 40.4           & 34.4      & 46.1           & 55.2       \\
        HARP (DW) & 40.6           & 78.2      & 46.6           & 1727.7    \\
        Node2Vec  & 40.5 & 78.2 & 45.5 & 951.2 \\
        MILE (NV) & {\bf 40.7}           & 50.5       & 46.3          & 83.5        \\
        HARP (NV) & 40.5           & 101.1      &  {\bf 47.2 }         & 1981.3       \\
        
        \hline
        \hline
      \end{tabular}
      
    }
    % \vspace{1mm}
  \caption{Comparisons of MILE with HARP.}
  %\caption{\small Comparisons of MILE with HARP.}
  \label{tab:HARP}
\vspace{-1em}
\end{table}

\subsection{Comparing MILE with Pytorch-biggraph}

\begin{figure}[t]
	\centering
	\setcounter{subfigure}{0}
	\includegraphics[width=0.8\linewidth]{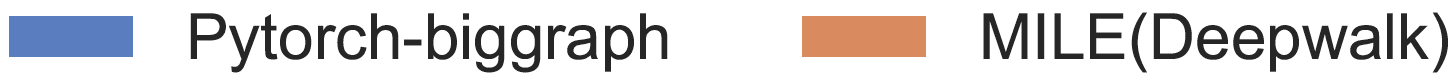}
	\subfloat[Node Classification]{\label{fig:bigraph_nc}\includegraphics[width=0.49\linewidth]{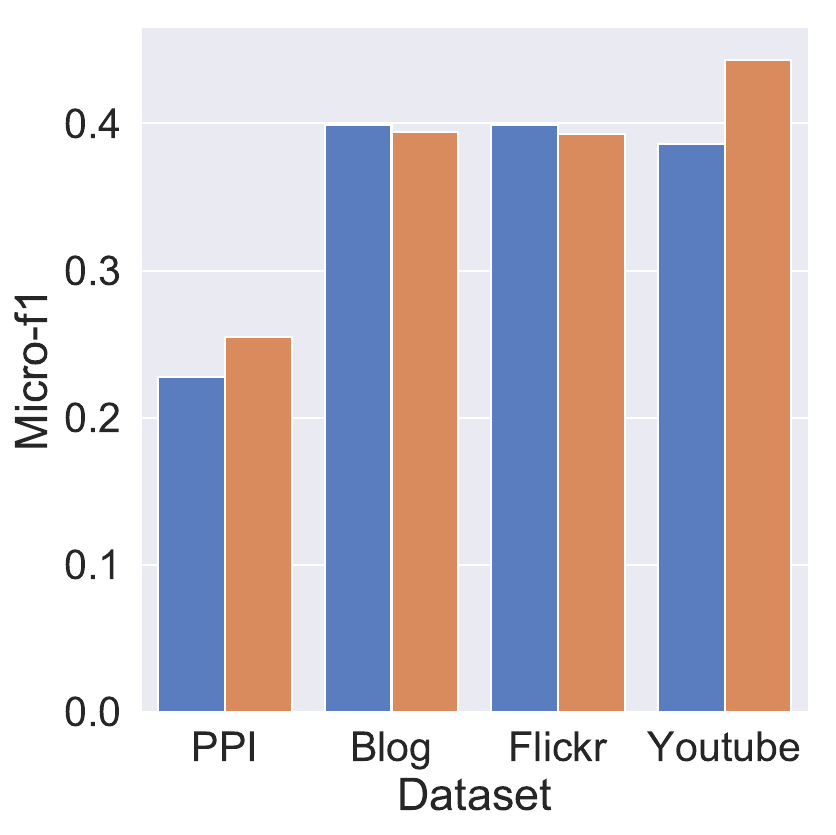}} %\hspace{4mm}
	\subfloat[Running time]{\label{fig:mem-nm}\includegraphics[width=0.48\linewidth]{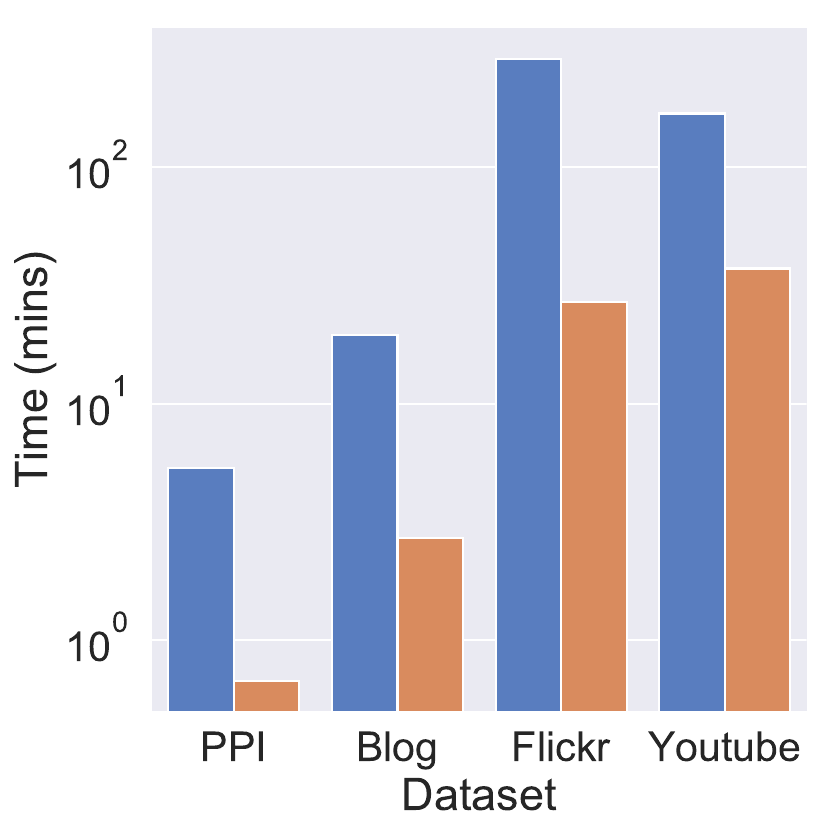}}
	\vspace{-0.5em}
	\caption{Node classification task:  Comparisons of MILE with Pytorch-biggraph.\hl{}}
	\label{fig:nc_biggraph}
	\vspace{-0.5em}
	
\end{figure}

\begin{figure}[t]
	\centering
	\setcounter{subfigure}{0}
	\includegraphics[width=0.8\linewidth]{Figures/biggraph.png}
	\subfloat[Link Prediction]{\label{fig:bigraph_nc}\includegraphics[width=0.49\linewidth]{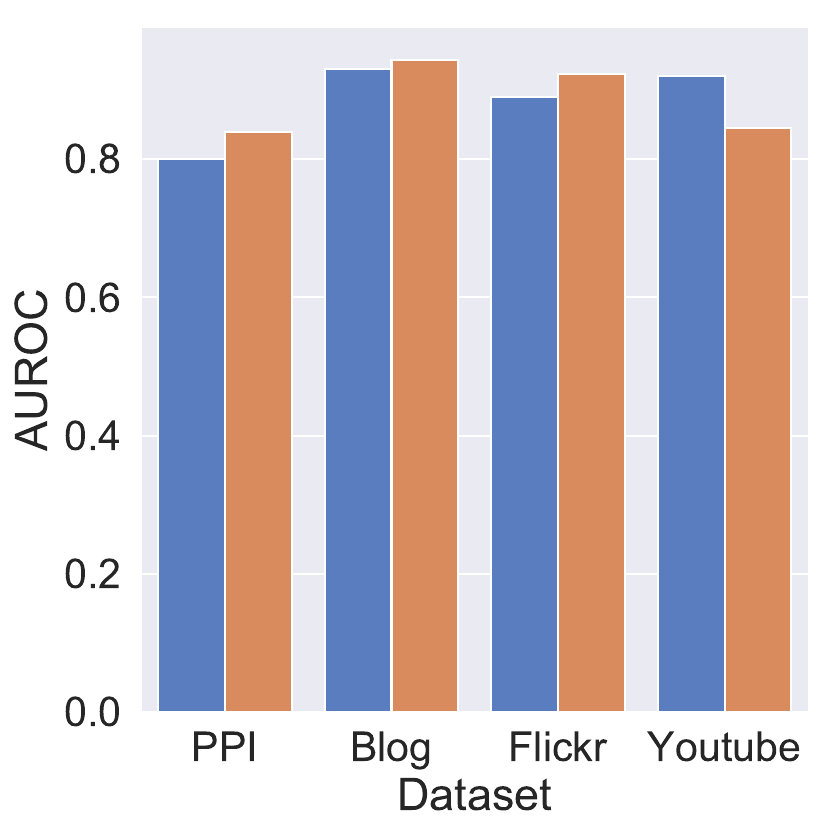}} %\hspace{4mm}
	\subfloat[Running time]{\label{fig:mem-nm}\includegraphics[width=0.48\linewidth]{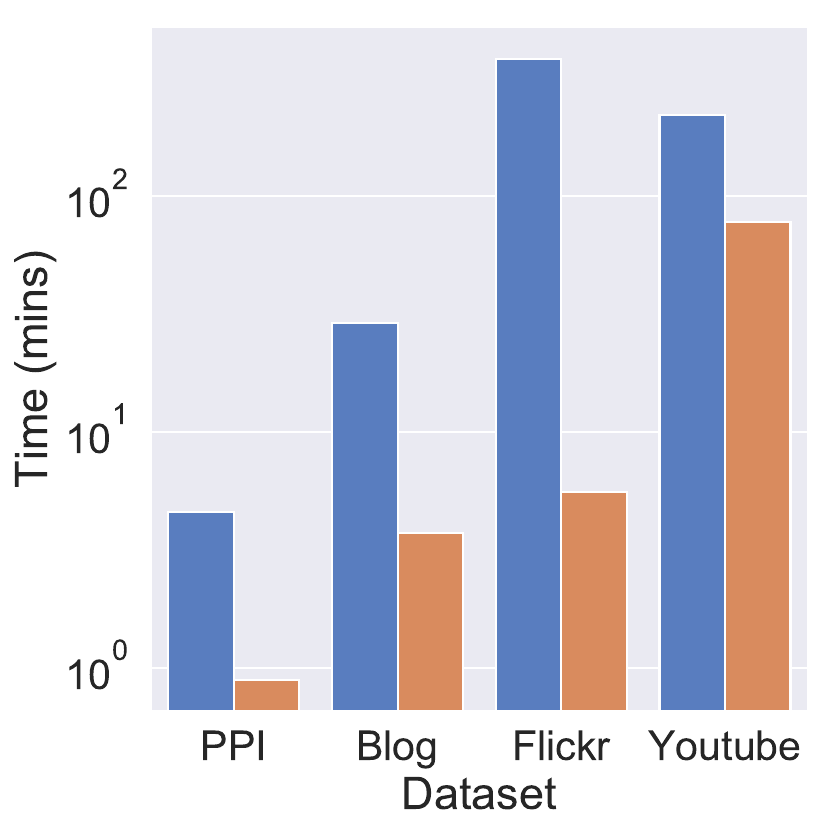}} %\hspace{-1mm}
	% \vspace{-3mm}
	\caption{
		Link Prediction task:  Comparisons of MILE with Pytorch-biggraph.\hl{}
	}
	\label{fig:lp_biggraph}
	\vspace{-1em}
\end{figure}

In this section, we compare the performance of MILE with Pytorch-biggraph. In Pytorch-biggraph, we set the number of workers to 28 (equal to the number of available cores). Figure \ref{fig:nc_biggraph} and Figure \ref{fig:lp_biggraph} respectively show the node classification and link prediction performance of both Pytorch-biggraph and MILE (Deepwalk) methods. Note that, in MILE, the speedup is driven by the coarsening level and higher speedup is achieved with higher coarsening level -- as evident from the experiments shown in Figure \ref{fig:level} and Figure \ref{fig:link_level}. For PPI and Blog dataset, we set coarsen level to 2 and for Flickr and YouTube dataset, we set  coarsen level to 6. From Figure \ref{fig:nc_biggraph}, we see that for the node classification task on PPI and YouTube dataset, MILE outperforms Pytorch-biggraph on classification quality metrics (higher-quality embeddings) while the running time of MILE is lower than Pytorch-biggraph. The lower running time of MILE is due to the proposed multi-level framework.   
From Figure \ref{fig:lp_biggraph}, we observe that MILE outperforms Pytorch-biggraph for the link prediction task on three datasets namely PPI, Blog and Flickr with less running time compared to Pytorch-biggraph. This experiment shows that on a single system with 28-cores, MILE outperforms Pytorch-biggraph in terms of running time and also outperforms Pytorch-biggraph on downstream tasks such as node classification and link prediction. 
\begin{leftbar}
\end{leftbar}

%\resizebox{0.8\columnwidth}{!}{
%}

\vspace{-1em}
\input{drilldown.tex}

%% file: drilldown.tex
\section{MILE Drilldown}

\subsection{Design Choices}
\label{subsect:drilldown:design}

%TODO(jiongqian): get rid of "this category of methods"; More rigorous on the levels.
We now study the role of the design choices we make within the MILE framework related to the coarsening and refinement procedures
described. 
To this end, we examine alternative design choices and systematically examine their performance.
The alternatives we consider are:
% \begin{itemize}[leftmargin=*]

\smallskip
\noindent \textbf{Random Matching (MILE-\texttt{rm})}: 
%We replace Algorithm~\ref{algo:coarsen} with a simple random matching approach for graph coarsening. 
For each iteration of coarsening, we repeatedly pick a random pair of connected nodes as a match and merge them into a super-node until no more matching can be found. The rest of the algorithm is the same as our MILE. %Using DeepWalk and NetMF for base embedding, we denote this category of variant as MILE-DW-\texttt{rm} and MILE-NM-\texttt{rm} respectively.

\smallskip
\noindent \textbf{Simple Projection (MILE-\texttt{proj})}: We replace our embedding refinement model with a simple projection method. In other words, we directly copy the embedding of a super-node to its original node(s) without any refinement ( $\mathcal{E}^p_{i} = M_{i, i+1} \mathcal{E}_{i+1}$ ).

%(see Eq.~\ref{eq:proj}). %We denote this categories of method as MILE-DW-\texttt{proj} and MILE-NM-\texttt{proj}.

\smallskip
\noindent \textbf{Averaging Neighborhoods (MILE-\texttt{avg})}: For this baseline method, the refined embedding of each node is a weighted average node embedding of its neighborhoods (weighted by the edge weights). This can be regarded as an embeddings propagation method. We add self-loop to each node\footnote{Self-loop weights are tuned to the best performance.} and conduct the embeddings propagation for two rounds.% (each node updates its embedding twice). %We represent the baselines as MILE-DW-\texttt{avg} and MILE-NM-\texttt{avg}.

\smallskip
\noindent \textbf{Untrained Refinement Model (MILE-\texttt{untr})}: Instead of training the refinement model to minimize the loss defined %in Eq.~\ref{eq:loss2}, 
as $L = \frac{1}{|V_m|} \left\Vert \mathcal{E}_{m} - H^{(l)}(\mathcal{E}_{m}, A_m) \right\Vert^2$, 
this baseline merely uses a fixed set of values for parameters $\Theta^{(k)}$ without training (values are randomly generated; other parts of the refinement model %Eq.~\ref{eq:layer} 
are the same, including $\tilde{A}$ and $\tilde{D}$). 
%We denote the baselines as MILE-DW-\texttt{untr} and MILE-NM-\texttt{untr}.

\smallskip
\noindent \textbf{Double-base Embedding for Refinement Training (MILE-\texttt{2base})}: This method replaces the loss function %as Eq.~\ref{eq:loss2} 
$L = \frac{1}{|V_m|} \left\Vert \mathcal{E}_{m} - H^{(l)}(\mathcal{E}_{m}, A_m) \right\Vert^2$
with %the alternative one in Eq.~\ref{eq:loss1} 
$L = \frac{1}{|V_m|} \left\Vert \mathcal{E}_{m} - H^{(l)}(M_{m, m+1} \mathcal{E}_{m+1}, A_m) \right\Vert^2$
for model training. 
It conducts one more layer of coarsening and 
base embedding (level $m+1$), from which the embeddings are projected to level $m$ and used as the input for model training. 

\smallskip
\noindent \textbf{GraphSAGE as Refinement Model (MILE-\texttt{gs})}: It replaces the graph convolution network in our refinement step with GraphSAGE~\cite{GRAPHSAGE}\footnote{Adapt code from \url{https://github.com/williamleif/GraphSAGE}}. 
We choose max-pooling for aggregation and set the number of sampled neighbors as $100$, as suggested by the authors. Also, concatenation is conducted instead of replacement during the process of propagation. 
% \end{itemize}
\input{table2.tex}

\subsection{Results of MILE-variants}

Table~\ref{tab2} shows the comparison of performance on these methods across the four datasets. 
%Due to the limit of space, 
Here, we focus on using DeepWalk and NetMF for base embedding with coarsening level as $m=1$ for PPI, Blog, and Flickr, while $m=6$ for YouTube. Results are similar for the other embedding options we consider. 
% The results on Node2Vec and GraRep as well as the ones with larger levels are similar.
%Note that we cannot apply HARP on NetMF since NetMF does not take initialized values to solve the matrix factorization by default.
%Though it is possible to solve the matrix factorization using Stochastic Gradient Descent method, it is out-of-scope in the current effort and we do not run HARP on NetMF. 
We hereby summarize the key information derived from Table~\ref{tab2} as follows:
% \begin{itemize}
% \vspace{-2mm}

\smallskip
\noindent {\bf The matching methods used within MILE offer a qualitative benefit at a minimal cost to execution time.} Comparing 
MILE with MILE-\texttt{rm} for all the datasets, we can see that MILE generates better embeddings than MILE-\texttt{rm} using either DeepWalk or NetMF as the base embedding method. 
Though MILE-\texttt{rm} is slightly faster than MILE due to its random matching, 
its Micro-F1 score are consistently lower than of MILE. 
% and Macro-F1 score

\smallskip
\noindent  {\bf The graph convolution based refinement learning methodology in MILE is particularly effective.} 
Simple projection-based MILE-\texttt{proj}, performs significantly worse than MILE.
The other two variants (MILE-\texttt{avg} and MILE-\texttt{untr}) which do not train the refinement model at all, also perform much worse than the proposed method. Note MILE-\texttt{untr} is the same as MILE except it uses a default set of parameters instead of learning those parameters. Clearly, the model learning part of our refinement method is a fundamental contributing factor to the effectiveness of MILE.
Through training, the refinement model is tailored to the specific graph under the base embedding method in use.
The overhead cost of this learning (comparing MILE with MILE-\texttt{untr}),
can vary depending on the base embedding employed (for instance on the YouTube dataset, it is an insignificant 1.2\% on DeepWalk - while being up to 20\% on NetMF) but is still
worth it due to qualitative benefits (Micro-F1 up from 30.2 to 40.9 with NetMF on YouTube).

\smallskip
\noindent {\bf Graph convolution refinement learning outperforms GraphSAGE.}
Replacing the graph convolution network with GraphSAGE for embedding refinement,
MILE-\texttt{gs} does not perform as well as MILE. It is also computationally more expensive, partially 
due to its reliance on embeddings concatenation, instead of replacement, during the process the embeddings 
propagation (higher model complexity). 
% the other reason is it samples the same number of neighbors for all the nodes.

\smallskip
\noindent  {\bf Double-base embedding learning is not effective.}
In the section -- Intricacies of Refinement Learning -- we discussed the issues with unaligned embeddings
%arising from
of
the double-base embedding
method for the refinement model learning. 
The performance gap between MILE and MILE-\texttt{2base} in Table~\ref{tab2} provides empirical evidence 
supporting our argument.        
This gap is likely caused by the fact that the base embeddings of level $m$ and level $m+1$ might not lie in the same embedding space (rotated by some orthogonal matrix)~\cite{GRAPHSAGE}. 
As a result, using the projected embeddings $\mathcal{E}^p_m$ as input for model training (MILE-\texttt{2base}) is not as good as directly using $\mathcal{E}_m$ (MILE). Moreover, Table~\ref{tab2} shows that the additional round of base embedding in MILE-\texttt{2base} introduces a non-trivial overhead. On YouTube, the running time of MILE-\texttt{2base} is $1.6$ times as much as MILE.

%% file: table2.tex
\begin{table}[t]
\centering
\resizebox{1.0\linewidth}{!}{
\begin{tabular}{l|l|l|l|l|l|l|l|l}
\hline \hline
                             & \multicolumn{2}{c|}{PPI}                                  & \multicolumn{2}{c|}{Blog}                                 & \multicolumn{2}{c|}{Flickr}                               & \multicolumn{2}{c}{YouTube}                             \\ \hline
                             & \multicolumn{1}{c|}{Mi-F1} & \multicolumn{1}{c|}{Time} & \multicolumn{1}{c|}{Mi-F1} & \multicolumn{1}{c|}{Time} & \multicolumn{1}{c|}{Mi-F1} & \multicolumn{1}{c|}{Time} & \multicolumn{1}{c|}{Mi-F1} & \multicolumn{1}{c}{Time} \\ \hline
DeepWalk                     & 23.0                          & 2.42                      & 37.0                          & 8.02                      & 40.0                          & 50.08                     & 45.2                          & 604.83                   \\
MILE (DW)                    & \textbf{25.6}                 & 1.22                      & \textbf{42.9}                 & 4.69                      & \textbf{40.4}        & 34.48                     & \textbf{46.1}        & 55.20                    \\
MILE-\texttt{rm} (DW)      & 25.3        & 1.01                      & 40.4                          & 3.62                      & 38.9                          & 26.67                     & 44.9                          & 55.10                    \\
MILE-\texttt{proj} (DW)    & 20.9                          & 1.12                      & 34.5                          & 3.92                      & 35.5                          & 25.99                     & 40.7                          & 53.97                    \\
MILE-\texttt{avg} (DW)     & 23.5                          & 1.07                      & 37.7                          & 3.86                      & 37.2                          & 25.99                     & 41.4                          & 55.26                    \\
MILE-\texttt{untr} (DW) & 23.5                          & 1.08                      & 35.5                          & 3.96                      & 37.6                          & 26.02                     & 41.8                          & 54.52                    \\
MILE-\texttt{2base} (DW)   & 25.4                          & 2.22                      & 35.6                          & 6.74                      & 37.7                          & 53.32                     & 41.6                          & 94.74                    \\
MILE-\texttt{gs} (DW)      & 22.4                          & 2.03                      & 35.3                          & 6.44                      & 36.4                          & 44.81                     & 43.6                          & 394.72                   \\ \hline \hline
% HARP (DW)                    & 24.1                          & 3.08                      & \textit{41.3}        & 9.85                      & \textbf{40.6}                 & 78.21                     & \textbf{46.6}                 & 1727.78                  \\ \hline \hline
NetMF                        & 24.6                          & 0.65                      & 41.4                          & 2.64                      & 31.8                          & 69.72                     & N/A                           & $>$574                \\
MILE (NM)                    & \textbf{26.9}                 & 0.27                      & \textbf{43.8}                 & 1.98                      & \textbf{39.3}                 & 24.03                     & \textbf{40.9}                 & 35.22                    \\
MILE-\texttt{rm} (NM)      & 25.2                          & 0.22                      & 41.0                          & 1.69                      & 37.6                          & 20.00                     & 39.6                          & 33.52                    \\
MILE-\texttt{proj} (NM)    & 23.5                          & 0.12                      & 38.7                          & 1.06                      & 34.5                          & 15.10                     & 26.4                          & 26.48                    \\
MILE-\texttt{avg} (NM)     & 24.5                          & 0.13                      & 39.9                          & 1.05                      & 36.4                          & 14.86                     & 26.4                          & 27.71                    \\
MILE-\texttt{untr} (NM) & 24.8                          & 0.13                      & 39.4                          & 1.08                      & 36.4                          & 15.23                     & 30.2                          & 27.20                    \\
MILE-\texttt{2base} (NM)   & 26.6                          & 0.29                      & 41.3                          & 2.33                      & 37.7                          & 31.65                     & 34.7                          & 55.18                    \\
MILE-\texttt{gs} (NM)      & 24.8                          & 1.08                      & 40.0                          & 3.70                      & 35.1                          & 34.25                     & 36.4                          & 345.28                   \\ \hline \hline
\end{tabular}
}
% \vspace{1m}
\caption{
%\small
Comparisons of graph embeddings between MILE and its variants. 
Except for the original methods, the number of coarsening level $m$ is set to $1$ on PPI/Blog/Flickr and $6$ on YouTube. Mi-F1 is the Micro-F1 score in $10^{-2}$ scale while Time column shows the running time of the method in minutes. 
% The methods with top-2 micro-F1 scores are bolded and italicized respectively.
``N/A'' denotes the method consumes more than 128 GB RAM.}
\label{tab2}
%\extvspace{-4mm}
\vspace{-1em}
\end{table}

%% file: conclusion.tex
% \vspace{-2mm}
\normalsize
In this work, we propose a novel multi-level embedding (MILE) framework to scale up graph embedding techniques,
without modifying them. Our framework incorporates existing embedding techniques as black boxes, and significantly improves the scalability of extant 
methods by reducing both the running time and memory consumption. %, making many methods applicable to large-scale datasets.  
Additionally, MILE also provides a lift in the quality of node embeddings in 
most of the cases. A fundamental contribution of MILE is its ability to learn a refinement strategy that depends on both the underlying graph properties and the embedding method in use. \looseness=-1

\ifext
\vspace{2mm}
\noindent

\textbf{Acknowledgments}:
This work is supported by the National Science Foundation
under grants EAR-1520870 and DMS-1418265. Computational support was provided by
the Ohio Supercomputer Center under grant PAS0166.
All content represents the opinion of the authors,
which is not necessarily shared or endorsed by their sponsors.
\fi